\documentclass[twoside,11pt]{article}

\usepackage{jmlr2e}
\usepackage{amsmath}
\usepackage{algorithm}
\usepackage{algorithmic}
\usepackage{graphicx}
\usepackage{epstopdf}
\usepackage{caption}
\usepackage{subcaption}

\newcommand{\R}{\mathbb{R}}
\renewcommand{\P}{\statistic{P}}

\newcommand{\norm}[1]{\|#1\|}

\newcommand{\mat}[1]{\mathbf{#1}}

\newcommand{\B}{\mat{B}}
\newcommand{\M}{\mat{M}}
\newcommand{\N}{\mat{N}}

\newcommand{\Z}{\mat{Z}}
\newcommand{\J}{\mat{J}}
\newcommand{\Q}{\mat{Q}}
\newcommand{\I}{\mat{I}}

\renewcommand{\P}{\mat{P}}

\newcommand{\Usv}{\mat{U}}

\newcommand{\bone}{\boldsymbol{\beta}_0}
\newcommand{\binf}{\boldsymbol{\beta}_{\infty}}
\newcommand{\mone}{\mathbf{m}_0}
\newcommand{\SDM}{\mathbf{H}}

\newcommand{\PP}{\mathbb{P}}
\newcommand{\Acts}{\mathcal{A}}
\newcommand{\Obs}{\mathcal{O}}

\newcommand{\minf}{\bold{m}_{\infty}}
\newcommand{\Mao}{\bold{M}_{ao}}
\newcommand{\Cao}{\bold{C}_{ao}}
\newcommand{\cinf}{\bold{c}_{\infty}}
\newcommand{\Maoij}{\bold{M}_{a^lo^k}}
\newcommand{\Pth}{\boldsymbol{\mathcal{P}}_{\mathcal{T},\mathcal{H}}}
\newcommand{\Ph}{\boldsymbol{\mathcal{P}}_{\mathcal{H}}}
\newcommand{\Pthao}{\boldsymbol{\mathcal{P}}_{\mathcal{T}, \textrm{$ao$}, \mathcal{H}}}
\newcommand{\Pthe}{\hat{\boldsymbol{\mathcal{P}}}_{\mathcal{T},\mathcal{H}}}
\newcommand{\Phe}{\hat{\boldsymbol{\mathcal{P}}}_{\mathcal{H}}}
\newcommand{\Pthaoe}{\hat{\boldsymbol{\mathcal{P}}}_{\mathcal{T}, \textrm{$ao$}, \mathcal{H}}}
\newcommand{\TT}{\mathcal{T}}
\newcommand{\HH}{\mathcal{H}}
\newcommand{\PhiH}{\boldsymbol{\Phi}_{\HH}}
\newcommand{\PhiT}{\boldsymbol{\Phi}_{\TT}}
\newcommand{\ind}{\mathbb{I}}
\newcommand{\seq}{z}
\newcommand{\Seq}{Z}
\newcommand{\cPth}{\boldsymbol{\Sigma}_{\mathcal{T},\mathcal{H}}}

\newcommand{\cPthe}{\hat{\boldsymbol{\Sigma}}_{\mathcal{T},\mathcal{H}}}
\newcommand{\cPhe}{\hat{\boldsymbol{\Sigma}}_{\mathcal{H}}}

\newcommand{\QQ}{\mathcal{Q}}
\newcommand{\Bao}{\mathbf{B}_{ao}}

\newcommand{\Pqh }{\boldsymbol{\mathcal{    {P}}}_{\QQ,\HH}}
\newcommand{\ePqh}{\hat{\boldsymbol{\mathcal{P}}}_{\QQ,\HH}}
\newcommand{\Pqoh }{\boldsymbol{\mathcal{    {P}}}_{\QQ,ao,\HH}}
\newcommand{\ePqoh}{\boldsymbol{\mathcal{\hat{P}}}_{\QQ,ao,\HH}}
\newcommand{\Phh }{\boldsymbol{\mathcal{    {P}}}_{h}}
\newcommand{\Pqhh }{\boldsymbol{\mathcal{    {P}}}_{\QQ,h}}
\newcommand{\Pqohh }{\boldsymbol{\mathcal{    {P}}}_{\QQ,ao,h}}
\newcommand{\ePhh }{\hat{\boldsymbol{\mathcal{P}}}_{h}}
\newcommand{\ePqhh }{\hat{\boldsymbol{\mathcal{\P}}}_{\QQ,h}}
\newcommand{\ePqohh }{\hat{\boldsymbol{\mathcal{P}}}_{\QQ,ao,h}}
\newcommand{\mb}{\mathbf}
\newcommand{\UU}{\hat{\mb{U}}}
\newcommand{\VV}{\hat{\mb{V}}}
\newcommand{\Sing}{\hat{\mb{S}}}

\usepackage{color}

\ShortHeadings{Compressed Predictive States}{Hamilton, Milani Fard and Pineau}
\firstpageno{1}

\begin{document}

\title{Efficient Learning and Planning with Compressed Predictive States}

\author{\name William Hamilton  \email william.hamilton2@mail.mcgill.ca 
       \\ 
 	   {\name Mahdi Milani Fard \email mmilani1@cs.mcgill.ca }
   		\\
 	   {\name Joelle Pineau \email jpineau@cs.mcgill.ca} \\
       \addr School of Computer Science\\
       McGill University\\
       Montreal, QC, Canada
}

\editor{}

\maketitle

\begin{abstract}%   <- trailing '%' for backward compatibility of .sty file
Predictive state representations (PSRs) offer an expressive framework for modelling partially observable systems.
By compactly representing systems as functions of observable quantities, the PSR learning approach avoids using local-minima prone expectation-maximization and instead employs a globally optimal moment-based algorithm.
Moreover, since PSRs do not require a predetermined latent state structure as an input, they offer an attractive framework for model-based reinforcement learning when agents must plan without a priori access to a system model.
Unfortunately, the expressiveness of PSRs comes with significant computational cost, and this cost is a major factor inhibiting the use of PSRs in applications. 
In order to alleviate this shortcoming, we introduce the notion of compressed PSRs  (CPSRs). The CPSR learning approach combines recent advancements in dimensionality reduction, incremental matrix decomposition, and compressed sensing.
We show how this approach provides a principled avenue for learning accurate approximations of PSRs, drastically reducing the computational costs associated with learning while also providing effective regularization. 
Going further, we propose a planning framework which exploits these learned models.
And we show that this approach facilitates model-learning and planning in large complex partially observable domains, a task that is infeasible without the principled use of compression.\footnote{An earlier version of this work appeared as: W.L. Hamilton, M. M. Fard, and J. Pineau. Modelling sparse dynamical systems with compressed predictive state representations. In \emph{Proceedings of the Thirtieth International Conference on Machine Learning}, 2013.}

\end{abstract}


\section{Introduction}

In the reinforcement learning (RL) paradigm, an agent in a system acts, observes, and receives feedback in the form of numerical signals \citep{Sutton:1998}.
Given this experience, the agent determines an optimal policy (i.e., a guide for its future actions) via value-function based dynamic programming or parametrized policy search.
This is conceptually analogous to the `operant conditioning' postulated to underlie certain forms of animal (and human) learning.
Organisms learn to repeat actions that give positive feedback and avoid those with negative results.

\subsection{Fully to Partially Observable Domains}

In the standard formulation, an RL agent is given prior knowledge of a domain in the form of a state-space, transition probabilities, and an observation (i.e., sensor) model. 
Formally, the system is described by a Markov decision process (MDP), and given the MDP description, a variety of optimization algorithms may then be used to solve the problem of determining an optimal action policy \citep{Sutton:1998}.
In general, approximate solutions are determined for domains exhibiting large, or even moderate, dimensionality \citep{Gordon:1999}.

The situation is further complicated in domains exhibiting partial observability, where observations are aliased and do not fully determine an agent's state in a system.
For example, an agent's sensors may indicate the presence of nearby objects but not the agent's global position within an environment.
To accommodate this uncertainty, the MDP framework is extended as partially observable Markov decision processes (POMDPs) \citep{Kaelbling:1998}.
Here, the true state is not known with certainty, and optimization algorithms must act upon belief states (i.e., probability distributions over the state-space).

\subsection{Model-Learning Before Planning}

The POMDP extension introduces a measure of uncertainty in the reinforcement learning paradigm.
Nevertheless, an agent learning a policy via the POMDP framework has access to considerable a priori knowledge:
Most centrally, the agent (which necessarily and implicitly contains the POMDP solver) has access to a description of the system in the form of an explicit state-space representation.
Moreover, in a majority of instances, the agent knows the probabilities governing the transitions between states, the observation functions governing the emission of observable quantities from these states, and the reward function specifying some empirical measure of ``goodness" for each state \citep{Kaelbling:1998}.

Access to such knowledge allows for the construction of optimal (or near-optimal) plans and is useful for real-world applications where considerable domain-specific knowledge is available.
However, the converse situation, where a (near)-complete system model is not known a priori, is both important and lags behind in terms of research results.
In such a setting, an agent must learn a system model prior to (or while simultaneously) learning an action policy. 
%For convenience, we refer to such learning as \emph{agnostic} (in the non-theological sense), as the learning agent makes no initial presumptions about the nature of the domain in which it is acting.

At an application level, there are many situations in which expert knowledge is sparse, and it is possible that even application domains with domain-knowledge could benefit from the use of algorithms that learn system models prior to planning and that are thus free from unintended biases introduced via expert-specified system models.
At a more theoretical level, the development of general agents that both learn system models and plan using such models is fundamental in the pursuit of creating truly intelligent artificial agents that can learn and succeed independent of prior domain knowledge.
%There are countless demonstrations of the proficiencies of domain-specific non-agnostic learning algorithms: given neat, pre-determined representations, artificial agents are able to perform at the state-of-the-art on challenging problems, such as navigating robotic helicopters or playing complex games (e.g., chess or PacMan) \citep{Ng:2000,Silver:2010}.
%Yet there is a paucity of results demonstrating artificial agents capable of solving such challenges using only execution traces (i.e., action-observation trajectories sampled through exploration). 

\subsection{Learning a Model-based Predictive Agent}

In this work we outline an algorithm for constructing a learning and planning agent for sequential decision-making under partial state observability.
At a high-level, the algorithm is model-based, specifying an agent that builds a model of its environment through experience and then plans using this learned model.
Such a model-based approach is necessary in complicated partially observable domains, where single observations are far from sufficient statistics for the state of the system \citep{Kaelbling:1998}. 
At its core, the algorithm relies on the powerful and expressive model class of predictive state representations (PSRs) \citep{Littman:2002}.
PSRs (described in detail in Section \ref{sec:psrs}) are an ideal candidate for the construction of an agent that both learns a system model and plans using this model, as they do not require a predetermined state-space as an input. 

PSRs have been used as the basis of model-based reinforcement learning agents in a number of recent works \citep{Boots:2009,Rosencrantz:2004,Ong:2012,Izadi:2008,James:2004}.
However, for these previous approaches, the time and space complexities of learning scale super-linearly in the maximum length of the trajectories used (see Section \ref{sec:cpsrs}).
In this work we use an approach that simultaneously ameliorates the efficiency concerns related to constructing PSRs and alleviates the need for domain-specific feature construction.
The model-learning algorithm, termed compressed predictive state representation (CPSR), uses random projections in order to efficiently learn accurate approximations of PSRs in sparse systems.
In addition, the approach utilizes recent advancements in incrementally learning transformed PSRs (TPSRs), providing further optimization \citep{Boots:2011}.
The details of the model-learning algorithm are provided in Section \ref{sec:learningcpsrs}. Section \ref{sec:theory} presents theoretical results pertaining to the accuracy of the approximate learned model and elucidates how our approach regularizes the learned model, trading off reduced variance for controlled bias.

The planning algorithm used is an extension of the fitted-$Q$ function approximation-based planning algorithm for fully observable systems \citep{Ernst:2005}.
This approach has been applied to PSRs previously with some success \citep{Ong:2012} and provides a strong alternative to point-based value iteration methods \citep{Izadi:2008}.
%In particular, the theoretical error-bounds on PSR point-based value iteration require that the PSR states lie on a belief (i.e., probability) simplex \citep{Izadi:2008} --- implicitly restricting the space of possible PSR states that can be used to those that are directly mappable to POMDP-style belief states --- the fitted-$Q$ approach has no such requirement.
The algorithm simply substitutes a predictive state for the observable MDP state in a fitted-$Q$ learning algorithm, and a function approximator  is used to learn an approximation of the $Q$-function for the system (i.e., the function mapping predictive states and actions to expected rewards).
The details of the planning approach are outlined in Section \ref{sec:planning}.
The main empirical contribution of this work is the application of this approach to domains and sample-sizes of complexity not previously feasible for PSRs.
Section \ref{sec:empirical} will highlight empirical results demonstrating the performance of the algorithm on some synthetic robot navigation domains and a difficult real-world application task based upon the ecological management of migratory bird species. 

This work builds upon the algorithm presented in \cite{Hamilton:2013}, extending it in a number of ways. 
Specifically, this work (1) permits a broader class of projection matrices, (2) includes optional compression of both histories and tests, (3) combines compressed sensing with incremental matrix decomposition to facilitate incremental/online learning, (4) provides a more detailed theoretical analysis of the model-learning algorithm, (5) explicitly includes a planning framework, which exploits the learned CPSR models in a principled manner, and (6) provides extensive empirical results pertaining to both model-learning and planning, including results on a difficult real-world problem.

\section{Predictive State Representations}\label{sec:psrs}

Predictive state representations (PSRs) offer an expressive and powerful framework for modelling dynamical systems and thus provide a suitable foundation for a model-based reinforcement learning agent.
In the PSR framework, a predictive model is constructed directly from execution traces,  utilizing minimal prior information about the domain \citep{Littman:2002,Singh:2004}.
Unlike latent state based approaches, such as hidden Markov models or POMDPs, PSR states are defined only via observable quantities.
This not only makes PSRs more general, as they do not require a predetermined state-space, but it also increases their expressive power relative to latent state based approaches \citep{Littman:2002}.
In fact, the PSR paradigm subsumes POMDPs as a special case \citep{Littman:2002}.
In addition, PSRs facilitate model-learning without the use of local-minima prone expectation-maximization (EM) and allow for the efficient construction of globally optimal models via a method-of-moments based algorithm \citep{James:2004}.
The following section outlines the foundations of the PSR approach and sets the stage for the presentation of compressed predictive state representations in Section \ref{sec:cpsrs} and our efficient learning algorithm in Section \ref{sec:learningcpsrs}.
Much of the PSR background material (e.g., the derivation of the PSR model in Sections \ref{sec:techfoundations} and \ref{sec:psrmodel}) expands upon the presentation in \cite{Boots:2009} and utilizes important results from that work.

\subsection{Notation}\label{sec:notation}

\subsubsection{Matrix Algebra Notation}\label{sec:matnotation}

Bold letters denote vectors $\mat{v} \in \R^d$ and matrices $\mat{M} \in
\R^{d_1 \times d_2}$.
Given a matrix $\mat{M}$, $\norm{\mat{M}}$ denotes its Frobenius norm.
$\mat{M}^\dagger$ is used to denote the Moore--Penrose pseudoinverse of $\mat{M}$.
Sometimes names are given to the columns and rows of a matrix using
ordered index sets $\mathcal{I}$ and $\mathcal{J}$. In this case,  $\mat{M} \in \R^{\mathcal{I} \times \mathcal{J}}$ denotes a matrix of
size $|\mathcal{I}| \times |\mathcal{J}|$ with rows indexed by $\mathcal{I}$ and
columns indexed by $\mathcal{J}$.
We then specify entries in a matrix (or tensor) using these indices and the bracket notation; e.g., $[\mat{M}]_{i,j}$ corresponds to the entry in the row indexed by $i \in \mathcal{I}$ and the column indexed $j \in \mathcal{J}$.
Rows or columns of a matrix are specified using this index notation and the $*$ symbol; e.g., $[\mat{M}]_{i,*}$,  denotes the $ith$ row of $\mat{M}$.
Finally, given $\mathcal{I}' \subset \mathcal{I}$ and $\mathcal{J}' \subset \mathcal{J}$ we define $[\mat{M}]_{\mathcal{I}', \mathcal{J}'}$ as the submatrix of $\mat{M}$ with rows and columns specified by the indices in $\mathcal{I}'$ and $\mathcal{J}'$, respectively.

\subsubsection{Probability Notation}

We  denote the probability of an event by $\PP(\cdot)$ and  use $|$ to denote the usual probabilistic conditioning.
To avoid excessive notation, when the $\PP(\cdot)$ operator is applied to a vector of events, it is understood as returning a vector of probabilities unless otherwise indicated (i.e., a single operator is used for single events and vectors of events). 

For clarity, we use $||$  to denote conditioning upon an agents policy (i.e., plan).
That is, $||$ denotes that we are conditioning upon the knowledge that the agent will ``intervene'' in a system by executing the specified actions. 

\subsection{Technical Foundations}\label{sec:techfoundations}

A PSR model represents a partially observable system's state as a probability distribution over future events.
More formally, we maintain a probability distribution over different sequences of possible future action-observation pairs.
Such sequences of possible future action-observations are termed \emph{tests} and denoted $\tau$.
For example, we could construct a test 
$\tau_i = [o^{k_1}_{t+1}, o^{k_2}_{t+2}, ..., o^{k_n}_{t+n} || a^{l_1}_{t+1}, a^{l_2}_{t+2}, ..., a^{l_n}_{t+n}]$, where notationally subscripts refer to time, superscripts identify particular actions or observations, and actions following the $||$ symbol denote that we are conditioning upon the agent ``intervening" by performing those specified actions at the specified times.
We can then say that such a test is \emph{executed} if the agent intervenes and takes the specified actions, and we say the test \emph{succeeded} if the observations received by the agent match those specified by the test.
Going further, we can define the probability of success for test $\tau_i$ as 

\begin{equation}
\PP(\tau_i) = \PP(o^{k_1}_{t+1}, o^{k_2}_{t+2}, ..., o^{k_n}_{t+n} || a^{l_1}_{t+1}, a^{l_2}_{t+2}, ..., a^{l_n}_{t+n}).
\end{equation}

Of course, we want to know more than just the unconditioned probabilities of success for each test.
A complete model of a dynamical system also requires knowing the success probabilities for each test conditioned on the agent's previous experience, or \emph{history}.
We denote such a history 
$h_j = [a^{l_0}_{0}o^{k_0}_{0}, a^{l_1}_{1}o^{k_1}_1...a^{l_{t}}_{t}o^{k_{t}}_{t}]$, where again subscripts denote time and superscripts identify particular actions or observations.
Importantly,  the $||$ symbol for intervention is absent from the definition of history, as the sequence of actions specified in a history are assumed to have already been executed.

Finally, given that an agent has performed some actions and received some observations, defining some history $h_j$, we compute

\begin{equation}
\PP(\tau^{\Obs}_i | h_j || \tau^{\Acts}_i),
\end{equation}
the probability of $\tau_i$ succeeding  conditioned upon the agent's current history in the system, where $\tau^{\Acts}_i$ and $\tau^{\Obs}_i$ denote the ordered lists of actions and observations, respectively, defined in $\tau_i$.

It is not difficult to see that a partially observable system is completely described by the conditional success probabilities of  all tests given all histories.
That is, if we have $\PP(\tau^{\Obs}_i | h_j || \tau^{\Acts}_i) \: \forall i \:\forall j$ then we trivially have all necessary information to characterize the dynamics of a system.
Of course, maintaining all such probabilities directly is infeasible, as there is a potentially infinite number of tests and histories (and at the very least an exorbitant number for any system of even moderate complexity) \citep{Littman:2002}.

Fortunately, it has been shown that it suffices to remember only the conditional probabilities for a (potentially) small \emph{core set} of tests, and the conditional probabilities for all other tests may be defined as linear functions of the conditional probabilities for the tests in this core set\footnote{In this work, the shortened phrase \emph{core set} is always to be interpreted as \emph{core set of tests}; that is, such sets always correspond to a set of tests. } \citep{Littman:2002}. 
More formally, we define the system dynamics matrix, $\SDM$, as the (potentially infinite size) matrix, where each row corresponds to a particular test (under some lexicographic ordering), each column to a particular history (under some lexicographic ordering), and a particular $[\SDM]_{i,j}$ entry to $\PP(\tau^{\Obs}_i | h_j || \tau^{\Acts}_i)$.
$\SDM$ simply organizes $\PP(\tau^{\Obs}_i | h_j ||\tau^{\Acts}_i), \: \forall i \:\forall j$ in a matrix structure.
In \cite{Littman:2002} and \cite{Singh:2004} it is shown that if $\SDM$  has rank $k$ then (1) $k$ corresponds to the rank of the partially observable system, as defined by \cite{Jaeger:2000} and (2) there exists a \emph{minimal core set} of size $k$ (i.e., the smallest core set of tests is of size $k$, though there may be larger core sets).  Thus, if $\SDM$ has rank $k$,  it suffices to remember conditional probabilities for only $k$ tests (those that are a part of the minimal core set), and the conditional probabilities for all other tests may be defined as \emph{linear} functions of the conditional probabilities for these tests.

The rank of $\SDM$ thus describes the complexity of a system. For example,  a system with $\textrm{rank}(\SDM)=k$ can not be modelled by a POMDP with less than $k$ states; though it may require more than $k$ POMDP states \citep{Singh:2004}. 
In contrast, a PSR can always (exactly) model a system with $\textrm{rank}(\SDM)=k$ using a minimal core set of exactly size $k$ \citep{Singh:2004}.
This demonstrates how PSRs can be more compact than POMDPS.

Thus, for a PSR, given a minimal core set $\QQ$ (i.e., $|\QQ| = \textrm{rank}(\SDM)$), we can compute the conditional probability of some test $\tau_i \notin \QQ$ as 
\begin{equation}
\PP(\tau^{\Obs}_i | h_j || \tau^{\Acts}_i) = \mathbf{r}^\top_{\tau_i}\PP(\QQ^{\Obs}| h_j || \QQ^{\Acts}),
\end{equation}
where $\mathbf{r}_{\tau_i}$ is a vector of weights and $\PP(\QQ^{\Obs} | h_j || \QQ^{\Acts})$ an ordered vector of conditional probabilities for each test in the minimal core set $q_i \in \QQ$.
Integral to this approach is the fact that restricting the model to linear functions of tests in the minimal core set does not preclude the modelling of non-linear systems, as the dynamics implicit in the probabilities may specify non-linear behaviours \citep{Littman:2002}.

Thus, given the functions mapping tests in the core set to all other tests, it suffices to maintain, at time $t$, only the vector $\mathbf{m}_t =\PP(\QQ^{\Obs} | h_t || \QQ^{\Acts})$, where $h_t$ is the history of the system at time $t$.
That is, it suffices to maintain only the vector of conditional probabilities for the tests in a core set (which is usually assumed to be minimal) .

\subsection{The PSR Model}\label{sec:psrmodel}

Formally, a PSR model of a system is defined by $\langle \Obs, \Acts, \QQ, \mathcal{F}, \mone\rangle$, where $\Obs$ and $\Acts$ define the possible observations and actions respectively, $\QQ$ is a minimal core set of tests, $\mathcal{F}$ defines a set of linear functions mapping success probabilities for tests in the minimal core set to the probabilities for all tests, and $\mone$ defines the initial state of the system (i.e., $\mone  = \PP(\QQ^{\Obs} || \QQ^{\Acts})$).
Since $\mathcal{F}$  contains only linear functions,  its elements can be specified as vectors of weights.
These vectors, in turn, are specified using a finite set of linear operators (i.e., matrices).
Specifically, we define a linear operator $\Maoij$ for each action-observation pair such that 

\begin{align}
\PP(o^{k}_{t+1} | h_{t} ||  a^{l}_{t+1}) &= \minf^\top\Maoij \PP(Q^{\Obs} | h_t  || Q^{\Acts}) \\
&= \minf^\top\Maoij \mathbf{m}_t,
\end{align}
where $\minf$ is a constant normalizer such that $\minf^\top\mathbf{m}_t = 1, \: \forall t$.

These operators map probabilities of tests in the specified  minimal core set to the probabilities for single action-observation pairs and may be recursively combined to generate the full set of linear functions in $\mathcal{F}$.
For instance, for the test\\ $\tau_i = [o^{k_1}_{t+1}, o^{k_2}_{t+2}, ..., o^{k_n}_{t+n}  || a^{l_1}_{t+1}, a^{l_2}_{t+2}, ..., a^{l_n}_{t+n}]$, we compute
\begin{align}\label{eq:prediction}
\PP(\tau^{\Obs}_i | h_t || \tau^{\Acts}_i) &= \mathbf{r}^\top_{\tau_i}\PP(Q^{\Obs} | h_t || Q^{\Acts})\\
&= \minf^\top \mathbf{M}_{a^{l_n}o^{k_n}}\cdots\mathbf{M}_{a^{l_2}o^{k_2}}\mathbf{M}_{a^{l_1}o^{k_1}}\mathbf{m}_t.
\end{align}
These operators can also be used to produce $n$-step predictions (i.e., the probability \\$\PP(o^{k}_{t+n} | h_{t} || a^{l}_{t+n})$ of seeing an observation, $o^{k}$, after taking action, $a^l$, $n$-steps in the future) by:
\begin{align}
\PP(o^{k}_{t+n} | h_{t} || a^{l}_{t+n})  &= \minf^\top\Maoij(\mathbf{M}_{\star})^{n-1}\mathbf{m}_{t},
\end{align}
where $\mathbf{M}_{\star} = \sum_{a^{l}o^{k} \in \Acts \times \Obs} \Maoij$ is a matrix that can be computed once and stored as a parameter for quick computation  \citep{Wiewiora:2007}.

Lastly, the operators provide a convenient method for updating the predictive state, defined by the prediction vector $\mb{m}_t$, as an agent tracks through a system and receives observations.
The prediction vector $\mathbf{m}_t$ is updated to $\mathbf{m}_{t+1}$ after an agent takes an action $a^l$ and receives observation $o^k$ using:
\begin{align}\label{eq:update}
\mathbf{m}_{t+1} &= \PP(\QQ^{\Obs} | h_{t+1} || \QQ^{\Acts})\\
				&= \PP(\QQ^{\Obs} | h_{t}a^lo^k || \QQ^{\Acts})\\
				&= \frac{\Maoij\mathbf{m}_t}{\minf^\top\Maoij\mathbf{m}_t}.
\end{align}

Together, the elements of $\langle\Obs, \Acts, \QQ, \mathcal{F}, \mone\rangle$ (where  $\mathcal{F}$ is understood to contain the linear operators described above and the normalizer) thus provide a succinct model of a system, which allows for the efficient computation of event probabilities and also facilitates conditioning upon observed histories.

\subsection{Learning PSRs}\label{sec:learningpsrs}

There is a considerable amount of literature describing different approaches to learning PSRs. 
We provide an overview of the standard approaches, as Section \ref{sec:learningcpsrs} describes, in detail, the efficient compressed learning approach we propose.\footnote{For a slightly more detailed discussion of existing PSR learning approaches see \cite{Wiewiora:2007}.}

In general, PSR learning approaches may be divided into two distinct classes: discovery-based and subspace-based.
In the discovery-based approach, a form of combinatorial search is used to discover the (minimal) core set of tests, and the PSR model is then computed in a straightforward manner given the explicit knowledge of $\QQ$ \citep{James:2004, James:2005}.
This method generates an exact PSR model.
However, the combinatorial search required to find $\QQ$ precludes the use of this approach in domains of even moderate cardinality.

Unlike the discovery-based approaches, subspace-based approaches obviate the need for determining $\QQ$ exactly \citep{Hsu:2008,Boots:2009, Rosencrantz:2004}.
Instead, subspace-identification techniques (e.g., spectral methods) are used in order to find a subspace that is a linear transformation of the subspace defined by $\QQ$ \citep{Rosencrantz:2004}.
The linear nature of the PSR model allows the use of this transformed PSR model in place of the exact PSR model without detriment.
Specifically, it can be shown that the probabilities obtained via such a transformed model are consistent with those obtained via the true model \citep{Boots:2009}.

Formally, one first specifies a large (non-minimal) core set of tests $\TT$ and a set of histories $\HH$.
Next, one defines two \emph{observable matrices} $\Pth$, $\Ph$, and $|\Acts| \times |\Obs|$ \emph{observable matrices} $\Pthao$ (one for each action-observation pair).
$\Pth$ is a $|\TT| \times |\HH|$ matrix which contains the joint probabilities of all specified tests and histories.
$\Ph$ is a $|\HH| \times 1$ vector containing the marginal probabilities of each possible history.
And each $\Pthao$ is a $|\TT| \times |\HH|$ matrix containing the the joint probabilities of all specified tests and histories where a particular action-observation pair (indicated by the subscript) is appended to the history \citep{Boots:2009}. 
These observable matrices can be viewed as submatrices of $\SDM$, the system dynamics matrix (e.g., $\Pth = [\SDM]_{\TT,\HH}$).
We also define matrices $\Pqh$ and $\Pqoh \: \forall ao \in \Acts \times \Obs$ analogously but with $\QQ$ replacing $\TT$ (e.g., $\Pqh = [\Pth]_{\QQ,*}$).

Under the assumption that the empty history occurs first in the lexicographic ordering of $\HH$, the discovery-based approach builds a PSR model by
\begin{align}\label{eq:discoverylearn}
&\mone =  [\Pqh]_{*,1} \\
&\minf^\top = \Ph^\top(\Pqh)^{\dagger},\\
&\M_{ao} = \Pqoh(\Pqh)^\dagger,
\end{align}
while the subspace-based approach builds a model by
\begin{align}\label{eq:subspacelearn}
&\bone =  [\Z\Pth]_{*,1} \\
&\binf^\top = \Ph^\top(\Z\Pth)^{\dagger},\label{eq:binflearn} \\
&\B_{ao} = \Z\Pthao(\Z\Pth)^\dagger,\label{eq:Baolearn}
\end{align}
where $\Z$ is the projection matrix defining the subspace used for learning, which satisfies certain conditions.
The conditions upon $\Z$ and the standard selection criterion for choosing it are elucidated in Section \ref{sec:transformed} below.

From these equations we see that PSR learning, in both the subspace and discovery paradigms, corresponds to a set of regression problems.
The psuedoinverses in \eqref{eq:discoverylearn}-\eqref{eq:Baolearn} corresponding to solutions to a set regression problems.
For example, in the learning of $\minf$ the columns of $\Pqh$  correspond to samples in the regression (i.e., each history is a sample), the rows to features (i.e., each test is a feature), and the regression targets are the entries of $\Ph$ (i.e., the marginal history vector).

In general, the complexity of the discovery-based learning approach is dominated by the combinatorial search for the set of core tests. 
In the worst case this search has time-complexity $O((|\Acts||\Obs|)^L)$, where $L$ is the max-length of a trajectory (i.e., execution trace) used to learn the model.
If the minimal core set of tests is provided as input, the discovery-based method has complexity $O(|\HH||\QQ|^2)$; however, the assumption that the minimal core set of tests is known is not realistic in practice.
In contrast, the subspace-based approach has time-complexity $O(|\HH|||\TT|d_{\Z})$, where $d_{\Z}$ is the column-dimension of $\Z$.
If the size of the minimal core set of tests is known (an unrealistic assumption) then $d_{\Z} = |\QQ|$.

\subsection{Transformed Representations}\label{sec:transformed}
 
PSR models learned via the subspace method are often referred to as transformed PSRs (TPSRs), since they learn a model that is an invertible transform of a standard PSR model.
More formally, given the set of linear parameters defining a PSR model and an invertible matrix $\mathbf{J}$, we can construct a TPSR by applying $\mathbf{J}$ as a linear operator to each parameter.
That is, we set $\bone = \mathbf{J}\mathbf{m}_0$, $\binf^\top=\minf^\top\mb{J}^{-1}$, and $\mathbf{B}_{ao} = \mathbf{J}\Mao\mathbf{J}^{-1} \:\: \forall ao \in \Acts \times \Obs$, and these new transformed matrices constitute the TPSR model \citep{Boots:2011}.
It is easy to see that the $\mathbf{J}$'s cancel out in the prediction equation \eqref{eq:prediction} and update equation \eqref{eq:update}.
Intuitively, TPSRs can be thought of as maintaining a predictive state upon an invertible linear transform of the state defined by the tests in the minimal core set.

In practice, the matrix $\J$ is determined by the projection matrix $\Z$, which is used during learning in the subspace-based paradigm.
To make the relationship between $\J$ and $\Z$ explicit, we define the following matrices:  $\mb{R} = (\mb{r}_{\tau_i},\mb{r}_{\tau_2},...,\mb{r}_{\tau_{|\TT|}})^\top \in \mathbb{R}^{\TT \times \QQ}$ , with each row $i$ corresponding to the linear function mapping the probabilities of tests in the minimal core set to the probability of test $\tau_i$ (i.e., the $\mb{r}_{\tau_i}$ as defined in \eqref{eq:prediction}); $\N = \textrm{diag}(\Ph) \in \R^{\HH \times \HH}$, with the marginal history probabilities along the diagonal; and, $\Q \in \R^{\QQ \times \HH} $, with each column $j$ equal to the expected probability vector for the tests in the minimal core set given that history $h_j$ has been observed (i.e., $[\Q]_{*,j} = \M_{h_j}\mone$).
These matrices can then be used to define a factorization of the observable matrices.
In particular, \cite{Boots:2009} show that
\begin{equation}\label{eq:pthfactor}
\Pth = \mb{R}\Q\N 
\end{equation}
and that 
\begin{equation}\label{eq:pthaofactor}
\Pthao = \mb{R}\Mao\Q\N
\end{equation}
holds for all $ao \in \Acts \times \Obs$.

Examining the equations for the different learning methods (i.e., \eqref{eq:discoverylearn} and \eqref{eq:subspacelearn}) and using the factorizations given in \eqref{eq:pthfactor} and \eqref{eq:pthaofactor}, we see first that for the discovery-based method, which learns a true untransformed PSR, we have that 
\begin{equation}
\Pqh = \I\Q\N,
\end{equation}
where $\I$ is the identity.
In this case the set of tests in $\Pqh$ is the minimal core set, and thus the core set mapping operator $\mat{R}$ is replaced by the identity.
Similarly, we have
\begin{equation}
\Pqoh= \I\Mao\Q\N.
\end{equation}
Thus for the discovery method
\begin{align}
\Pqoh(\Pqh)^\dagger  &=\Mao\Q\N(\Q\N)^\dagger\\
&=\Mao,
\end{align}
where we used the fact that $\Q\N$ is full column-rank by definition.
By contrast, for the subspace learning algorithm we have, assuming that $\Z\mb{R}$ has full row-rank, 
\begin{align}
\B_{ao} &= \Z\Pthao(\Z\Pth)^\dagger\\
&= \Z\mb{R}\Mao\Q\N(\Z\mb{R}\Q\N)^{\dagger}\\
&= \Z\mb{R}\Mao\Q\N(\Q\N)^{\dagger}(\Z\mb{R})^{\dagger}\\
&= \Z\mb{R}\Mao(\Z\mb{R})^{\dagger}, \label{eq:almosttpsr}
\end{align}
where we again used the fact that $\Q\N$ has full column-rank.
If we further assume that $\Z\mb{R}$ is invertible (i.e., is square in addition to being full row rank) then \eqref{eq:almosttpsr} simplifies to
\begin{equation}
\Z\mb{R}\Mao(\Z\mb{R})^{-1}.
\end{equation}
Similar results hold for $\binf$ and $\bone$, showing that the subspace learning method does, in fact, return TPSRs in the case where $\Z\mb{R}$ is invertible, and in this case we have a transformed representation with  $\J := \Z\mb{R}$.

The final piece of a TPSR is the specification of $\mat{Z}$, the projection matrix defining the subspace used during learning (and implicitly defining the transformation matrix $\J$).
We know from the above derivations that $\Z$ must be chosen such that $\Z\mb{R}$ is invertible.
The standard method for guaranteeing this is by choosing $\mathbf{Z}$ via spectral techniques; that is, $\Z$ is set to be $\Usv^\top$, the transpose of the matrix of right singular vectors (from the thin-SVD of $\Pth$) \citep{Boots:2009}.
%Choosing $\Z$ in this way also allows for the tuning of the  dimension of the final representation by removing least significant vectors from $\Usv$ \cite{Boots:2009}.

The TPSR approach can also be extended to work with features of tests and histories \citep{Boots:2009,Boots:2011} and/or kernelized to work in continuous domains \citep{Boots:2013}.
This is useful in cases where the observation space is too complex for standard tests to be used (i.e., when the observation space is structured or continuous).
When features of tests and histories are used, however, they are usually specified in a domain-specific manner \citep{Boots:2009}.
Some authors have also used randomized Fourier methods to efficiently approximate kernel-based feature selection \citep{Boots:2011}.
These methods are quite successful in continuous domains \citep{Boots:2009, Boots:2011,Boots:2013}.

In contrast, the benefit of the algorithm presented in Section \ref{sec:learningcpsrs} is that it implicitly performs general purpose feature selection (for discrete-domains) using random compression.
And this is especially useful in cases where it is difficult to know a sufficient set of features prior to training (e.g., in the case where the model is being learned incrementally).
Moreover, the motivation between the compression performed in this work and the above-mentioned feature-based techniques are disjoint in that the goal of this work is to provide compression for efficient learning whereas the above-mentioned feature-based learning strategies are motivated by the need to cope with continuous or structured observation spaces.
See Section \ref{sec:related} for further discussion on the relationship between this work and these alternative feature-based approaches.

\section{Compressed Predictive State Representations}\label{sec:cpsrs}

In this section, we describe our extension of PSRs, compressed predictive state representations (CPSRs).
The CPSR approach, at its core, combines the state-of-the-art in subspace PSR learning with recent advancements in compressed sensing.
This marriage provides an extremely efficient and principled approach for learning accurate transformed approximations of PSRs in complex systems, where learning a full PSR is simply intractable.
Section \ref{sec:ce} motivates the use of compressed sensing techniques in a PSR learning algorithm, and Section \ref{sec:learningcpsrs} describes the efficient CPSR learning approach we propose.

\subsection{Foundations: Compressed Estimation}\label{sec:ce}

Despite the fact that non-compressed subspace-based algorithms, such as TPSR, can specify a small dimension for a transformed space (e.g., by removing the least important singular vectors of $\Usv$ as in done in \cite{Rosencrantz:2004} and analyzed in \cite{kulesza2014low}), there are still a number of computational limitations.
To begin, TPSRs require that the $|\TT| \times |\HH|$ matrix,  $\Pth$, be estimated in its entirety, and that the $\Pthao$ matrices be partially estimated as well. 
Moreover, since the naive TPSR approach must compute a spectral decomposition of $\Pth$ it has computational complexity $O(|\HH||\TT|^2)$, in the batch (and incremental mini-batch) setting,  assuming the observable matrices are given as input.
Thus in domains that require many (possibly long) trajectories for learning or that have large observation spaces, such as those described in Section \ref{sec:empirical}, the naive TPSR approach becomes intractable, since $|\HH|$ and $|\TT|$ both scale as $O(L|Z|)$, where $L$ is the max length of a trajectory in a training set $Z$ of size $|Z|$.\footnote{Note that $|\HH|$ and $|\TT|$ scale linearly with the number of \emph{observed} test/histories. The $O(L|Z|)$ bound is thus pessimistic in that it assumes each training instance is unique.}\footnote{It is worth noting that no explicit bounds on the sample complexity of PSR learning have been elucidated. However, the sample complexity bounds of \cite{Hsu:2008} provide results for a special case of TPSR learning (i.e., no actions and only single length tests and histories). In general, PSR approaches are consistent estimators but cannot be assumed to be data efficient (thus emphasizing the need to accommodate large sample sizes).} 
In order to circumvent these computational constraints (and provide a form of regularization), the CPSR learning algorithm we propose (in the next section) performs \emph{compressed estimation}.

This method is borrowed from the field of compressed sensing and works by projecting matrices down to low-dimensional spaces determined via randomly generated bases.
More formally, a $m \times n$ matrix $\mathbf{Y}$ is compressed to a $d \times n$ matrix $\mathbf{X}$ (where $d<m$) by:
\begin{equation}\label{eq:matcompress}
\mathbf{X} = \mb{\Phi}\mathbf{Y},
\end{equation}
where $\mb{\Phi}$ is a $d \times m$ \emph{Johnson-Lindenstrauss matrix} (i.e., a matrix satisfying the Johnson-Lindenstrauss lemma) \citep{Baraniuk:2009}. 
Intuitively, a Johnson-Lindenstrauss matrix is a random matrix defining a low-dimensional embedding which approximately preserves Euclidean distances between projected points (i.e., the projection preserves the dot-product between vectors).
Different choices for $\mb{\Phi}$ are discussed in Section \ref{sec:empirical}.
It is worth noting that in our case, the matrix multiplication in \eqref{eq:matcompress} is in fact performed ``online", and the matrices corresponding to $\mb X$ and $\mb{\Phi}$ are never explicitly held in memory (details in Section \ref{sec:learningcpsrs}). 

The fidelity of this technique depends what is called the \emph{sparsity} of the matrix $\mathbf{Y}$.
Sparsity in this context refers to the maximum number of non-zero entries which occur in any column of $\mathbf{Y}$.
Formally, if we denote a column vector of $\mathbf{Y}$ by $\mathbf{y}_{i}$, we say that a matrix is $k$-sparse if:
\begin{equation}
k \geq  ||\mathbf{y}_{i}||_{0} \:\: \forall \mathbf{y}_{i} \in \mathbf{Y},
\end{equation}
where $||\cdot||_{0}$ denotes Donoho's zero ``norm'' (which simply counts the number of non-zero entries in a vector).

The technique is very well suited for application to PSRs.
Informally, the sparsity condition is the requirement that for every history $h_{j}$, only a subset of all tests have non-zero probabilities (a more formal definition appears in the theory section below).  This seems realistic in many domains.
For example, in the PocMan domain described below, we empirically found the average column sparsity of the matrices to be roughly 0.018\% (i.e., approximately 0.018\% of entries in a column were non-zero).
Moreover, as we will demonstrate empirically in Section \ref{sec:empirical}, certain noisy observation models induce sparsity that can be exploited by this approach.

\subsection{Efficiently Learning CPSRs}\label{sec:learningcpsrs}

In this section, we present our novel compressed predictive state representation (CPSR) learning algorithm.
The algorithm builds upon the work of \cite{Hamilton:2013}, extending their algorithm in a number of important ways.
Specifically, the algorithm presented here (1) permits a broad class of compression matrices (any full-rank projection matrix satisfying the JL lemma), (2) includes optional compression of both histories and tests, and (3) combines compressed sensing with spectral methods in order to provide numerical stability and facilitate incremental (and even online) model-learning. Section \ref{sec:batchlearn} describes the foundational batch-learning algorithm. Section \ref{sec:incupdates} describes how to incrementally update a learned model with new data efficiently for deployment in online settings.

\subsubsection{Batch Learning of CPSRs}\label{sec:batchlearn}

To begin, we define two injective functions: $\phi_\TT \::\: \TT \rightarrow \mathbb{R}^{d_\TT}$ and $\phi_\HH \::\: \HH \rightarrow \mathbb{R}^{d_\HH}$.
These functions are independent mappings from tests and histories, respectively, to columns of independent random full-rank Johnson-Lindenstrauss (JL) projection matrices $\PhiT \in \mathbb{R}^{d_\TT \times \TT}$ and $\PhiH \in \mathbb{R}^{d_\HH \times \HH}$, respectively.
The matrices are defined via these functions since the full sets $\TT$ and $\HH$ may not be known a priori, and we can get away with this ``lazy" specification since the columns of JL projection matrices are determined by independent random variables.

Next, given a training trajectory $\seq$ of action-observation pairs of any length, let $\ind_{h_j}(\seq)$ be an indicator function taking a value of $1$ if the action-observations pairs in $\seq$ correspond to $h_j$. 
Similarly define $|\cdot|$ as the length of a sequence (e.g., of action-observation pairs) and let $\ind_{h_j,\tau_i}(\seq)$ be an indicator function taking a value of $1$ if $\seq$ can be partitioned such that, starting from some index $k$ within the sequence, there are $|h_j|$ action-observation pairs corresponding to those in $h_j \in \HH$ and the next $|\tau_i|$ pairs correspond to those in $\tau_i \in \TT$.\footnote{In this work we use $k=0$. That is we do not use the suffix history estimation algorithm \citep{Wolfe:2005}, where $k$ is varied in the range $[0,|z|)$.
Using $k=0$ minimizes dependencies between estimation errors as the same samples are not used to get estimates for multiple histories.}

Given a batch of of training trajectories $\Seq$ we compute compressed estimates of the observable matrices $\Ph$ and $\Pth$\footnote{We do not normalize our probability estimates in the estimation equations since the normalization constants cancel out during learning.}:
\begin{align}\label{eq:chistest}
\cPhe &= \PhiH\Phe \\
	  &= \sum_{\seq \in \Seq}\sum_{h_j \in \HH}\ind_{h_j}(\seq)\phi_\HH(h_j),\\\nonumber\\
\cPthe &= \PhiT\Pthe\PhiH^\top\\ \label{eq:cpthest}
&= \sum_{\seq \in \Seq}\sum_{t_i,h_j \in \TT \times \HH} \ind_{h_j,t_i}(\seq) \left[\phi_\TT(t_i) \oplus \phi_\HH(h_j)\right],
\end{align}
where $\oplus$ denotes the tensor (outer) product of two vectors.

Next, we compute the $\UU\Sing\VV^\top$ rank-$d'$ thin SVD of $\cPthe$:
\begin{equation}\label{eq:svd}
(\UU,\Sing,\VV) = \textrm{SVD}(\cPthe).
\end{equation}
Given these matrices we can construct $\mathbf{c}_1$ and $\cinf$, the compressed and transformed estimates of $\mathbf{m}_1$ and $\minf$, respectively:
\begin{align}\label{eq:c1def}
\mathbf{c}_1 = \Sing\VV^\top\mathbf{e},
\end{align}
\begin{align}\label{eq:cinfdef}
\cinf^\top = \cPhe^\top\VV\Sing^{-1},
\end{align}
where $\mathbf{e}$ is a vector such that $\PhiH\mathbf{e} = (1,0,0,...,0)^\top$.
In practice this can be guaranteed by defining a modified history map $\phi'_\HH \::\: \HH \rightarrow \mathbb{R}^{d+1}$ such that that for the null history, $\emptyset$, $\phi'_\HH(\emptyset) = (1,0,0,...,0)^\top$ and that $\phi'_\HH(h_j) = [0 \:\: \phi_\HH(h_j)]$ for all $h_j \neq \emptyset$.
This specification of $\mathbf{e}$ assumes that all $\seq \in \Seq$ are starting from a unique start state.
If this is not the case, then we set $\mathbf{e}$ such that $\PhiH\mathbf{e} = (1,1,1,...,1)^\top$, which again can be guaranteed without cost but in this case by simply adding a constant ``dummy" column to the front of $\PhiH$.
In this latter scenario, we would, in fact, not be learning $\mathbf{c}_1$ exactly and instead would learn $\mathbf{c}_*$, an arbitrary feasible state as our start state.
The uncertainty in our state estimate should decrease, however, as we update and track through our system and the process mixes \citep{Boots:2009}.
And indeed, the majority of domains without well-defined start-states are those for which there is significant mixing over time, so this technique should introduce only a small amount of error in practice.

Given the SVD of $\cPthe$, we can also estimate the $\Cao$ matrices, the compressed and transformed versions of the $\Mao$ matrices, directly via a second pass over the data. 
First, however, we must define a third class of indicator functions on $z \in Z$: $\ind_{h_j,ao,\tau_i}(\seq)$ takes value $1$ if and only if the training sequence $\seq$ can be partitioned such that, starting from some index $k$ within the sequence, there are $|h_j|+1$ action-observation pairs corresponding to $h_j$ appended with a particular $ao \in \Acts \times \Obs$ and the next $|\tau_i|$ correspond to those in $\tau_i$.
In other words, $\ind_{h_j,ao,\tau_i}(\seq)$ is equivalent to $\ind_{h_j',\tau_i}(\seq)$, where a particular $ao \in \Acts \times \Obs$ is appended to the history $h_j'$.
Using these indicators and the SVD matrices of $\cPthe$, we compute, for each $ao \in \Acts \times \Obs$:

\begin{align}\label{eq:Caodef}
\Cao = \sum_{\seq \in \Seq}\sum_{t_i,h_j \in \TT \times \HH} \ind_{h_j,ao,t_i}(\seq)\left[ \left( \UU^\top\phi_\TT(t_i) \right) \oplus \left(\Sing^{-1}\VV^\top\phi_\HH(h_j)\right)\right].
\end{align}

Thus, in two passes over the data, we are able to efficiently construct our CPSR model parameters.
The primary computational savings engendered by this approach is in the computation of the pseudoinverse of $\cPthe$, which we implicitly compute via an SVD.
Since we are performing pseudoinversion (i.e., SVD) on a compressed matrix, the computational complexity is uncoupled from the number of tests and histories in the set of observed trajectories $Z$.
Recalling that $L$ denotes the max length of a trajectory in $Z$ and letting $|Z|$ denote the number of trajectories in the set $Z$, this approach has a computational complexity of 
\begin{equation}\label{eq:cpsrcomplex}
 O\left(L|Z|d_\HH d_\TT + d^2_\TT d_\HH\right) = O\left(L|Z|\right)
\end{equation}
since $d_\HH$ and $d_\TT$ are a user-specified constants\footnote{Section \ref{sec:theory} describes how the choice of these constants affects the accuracy of the learned model.} (assuming the standard cubic computational cost for the SVD).
Without compression (i.e., with naive TPSR), a computational cost of 
\begin{equation}
O\left(L|Z| + |\HH||\TT|^2\right) = O\left(L^3|Z|^3\right)
\end{equation}
is incurred.

In addition to these computational savings, the above approach has the added benefit of not requiring that $\TT$ and $\HH$ be known in entirety prior to learning.
This is especially important in the case where we want to alternate model-learning and planning/exploration phases using incremental updates (described below), as it is very unlikely that all possible tests and histories are observed in the first round of exploration.
Performing SVD on the compressed matrices also induces a form of regularization (similar to $L_2$ regularization) on the learned model, where variance is reduced at the cost of a controlled bias (details in Section \ref{sec:theory}).

\subsubsection{Incremental Updates to the Model}\label{sec:incupdates}

In addition to straightforward batch learning, it is also possible to incrementally update a learned model, given new training data, $\Seq'$ \citep{Boots:2011}.
This is especially useful in that it facilitates alternating exploration and exploitation phases.
Of course, if such a non-blind alternating approach is used then the distribution of the training data changes (i.e., it becomes non-stationary), and the sampled trajectories can no longer be assumed to be i.i.d..
Despite this theoretical drawback, \cite{Ong:2012} show that non-blind sampling approaches can lead to better planning results in a small sample setting.\footnote{In this work, where larger sample sizes were used, we did not find a significant benefit to goal-directed sampling and in fact saw detrimental effects in terms of planning ability and numerical stability during learning. See Section \ref{sec:discuss} for details.}

Briefly, we obtain a new $\cPthe$ estimate and update our $\cPhe$ estimate using using \eqref{eq:cpthest} and \eqref{eq:chistest} with $\Seq'$.
Next, we update our SVD matrices, given our additive update to $\cPthe$, using the methods of \citep{Brand:2002}.
The $\mathbf{c}_1$ and $\cinf$ vectors are then re-computed exactly as in equations \eqref{eq:c1def} and \eqref{eq:cinfdef}.

To obtain our $\Cao^{\textrm{new}}$ matrices, we compute:

\begin{align}\label{eq:newcao}
\Cao^{\textrm{new}} &= \sum_{\seq \in \Seq'}\sum_{t_i,h_j \in \TT \times \HH} \ind_{h_j,ao,t_i}(\seq)\left[ \left( \hat{\mb U}_{\textrm{new}}^\top\phi_\TT(t_i) \right) \oplus  \left(\hat{\mb S}^{-1}_{\textrm{new}}\hat{\mb V}_{\textrm{new}}^\top\phi_\HH(h_j)\right)\right] \nonumber \\
&+ \hat{\mb U}_{\textrm{new}}^\top\hat{\mb U}_{\textrm{old}}\Cao^{\textrm{old}}\hat{\mb S}_{\textrm{old}}\hat{\mb V}^\top_{\textrm{old}}\hat{\mb V}_{\textrm{new}}\hat{\mb S}^{-1}_{\textrm{new}}.
\end{align}
The first term in \eqref{eq:newcao} corresponds to estimating the contribution to the new $\Cao$ matrix from the new data, and the second term is the projection of the old $\Cao$ matrix onto the new basis. Using the results of \cite{Brand:2002}, the complexity of this update is \begin{equation}
O(L'|Z'|(d_\TT d_\HH + (d')^3 + d'd_\TT)+d_\TT d'd_\HH),
\end{equation}
 where $L'$ denotes the maximum length of a trajectory in $Z'$.

\section{Theoretical Analysis of the Learning Algorithm}\label{sec:theory}

In the following section, we describe theoretical properties of the CPSR learning approach.
Our analysis proceeds in two stages.
First, we show that the learned model is consistent in the case where $d_{\TT} \geq |\QQ|$ and $d_{\HH} \geq |\QQ|$ (i.e., when no real compression occurs).
Following this, we outline results bounding the induced approximation error (bias) and decrease in estimation error (variance) due to learning a compressed model.

The analysis included in this section is intended as a means to justify the compression technique and study the overall consistency of our algorithm. It also provides guidance for the choosing of a theoretically sound range of values for the projection size used in the algorithm.

\subsection{Consistency of the Learning Approach}\label{sec:consistency}

The following adapts the results of \cite{Boots:2009} and shows the consistency of our learning approach when the random projection dimension is greater than or equal to the true underling dimension of the system (i.e., the size of the minimal core set of tests, $|\QQ|$).
We then describe the implications of this result for the case where we are in fact projecting down to a dimension smaller than $|\QQ|$.

\subsubsection{Consistency in the Non-Compressed Setting}\label{sec:consistency}

We begin by noting a fundamental result from the TPSR literature.
Recall the matrix $\mb{R} = (\mb{r}_{\tau_1},\mb{r}_{\tau_2},...,\mb{r}_{\tau_{|\TT|}})^\top \in \mathbb{R}^{\TT \times \QQ}$ where each row, $\mb{r}_i$, specifies the linear map:
\begin{equation}
\mb{r}^\top_{\tau_i} \PP(Q^{\Obs} | h_t || Q^{\Acts})= \PP(\tau^{\Obs}_i | h_t || \tau_i^{\Acts}).
\end{equation}
Supposing that $d_{\TT} \geq |\QQ|$ and $d_{\HH} \geq |\QQ|$  and with $\Usv$ coming from the SVD of $\cPth$, we have
\begin{align}
\mb{c}_0 &= (\Usv^\top\PhiT\mat{R})\mb{m}_0,\label{eq:c1consist}\\
\cinf^\top &= \minf^\top(\Usv^\top\PhiT\mat{R})^{-1},\label{eq:cinfconsist}\\
\Cao &= (\Usv^\top\PhiT\mat{R})\Mao(\Usv^\top\PhiT\mat{R})^{-1}\label{eq:Caoconsist}.
\end{align}
That is, we simply recover a TPSR where $\mb{J} = (\Usv^\top\PhiT\mat{R})$, and it has been shown that the above implies a \emph{consistent} learning algorithm \citep{Boots:2009, Boots:2011}.
We note that $\PhiT$ appears in these consistency equations, while $\PhiH$ does not, emphasizing the different roles these two matrices occupy.
This difference will play an important role in the theoretical analysis below.

\subsubsection{Extension to the Compressed Case}\label{sec:extendcomp}

In the case where $d_{\TT} < |\QQ|$ and/or $d_{\HH} < |\QQ|$ things are not as straightforward.
Specifically, equations \eqref{eq:c1consist}-\eqref{eq:Caoconsist} no longer hold as $(\Usv^\top\PhiT\mat{R})$ is no longer invertible (it is in fact, no longer square), since the SVD is taken on $\cPth$ which has rank less than $|\QQ|$ when $d_\TT < |\QQ|$ and/or $d_\HH < |\QQ|$ while the column dimension of $\mb{R}$ is $|\QQ|$. 
The primary focus of our theoretical analysis is the effect of this fact, i.e. $(\Usv^\top\PhiT\mat{R})$ not being invertible.
We show how we can view $\PhiT$ as inducing a form of compressed linear regression, and we provide bounds on the excess risk of learning within a compressed space.

There is, however, the additional complication of $\PhiH$ when $d_{\HH} < |\QQ|$, as in that setting it is no longer possible to remove $\PhiH$ from the consistency equations \eqref{eq:c1consist}-\eqref{eq:Caoconsist}. 
From the perspective of regression, $\PhiH$ can be viewed as compressing the number of samples, while $\PhiT$ can be viewed as compressing the features. 
In this work, we focus on the effect of compressing tests and provide detailed analysis of how compressing tests (i.e., features) affects the implicit linear regression performed.
\cite{Zhou:2007} discuss the effect of compressing samples during regression, a result that follows naturally from the Johnson-Lindenstrauss lemma, and in Section \ref{sec:discuss}, we discuss these results and their relationship to this work.
For completeness, Section \ref{sec:empirical} also provides an empirical analysis of the effects of compressing histories and tests versus compressing tests alone.

\subsection{Effects of Compression}

In what follows, we analyse the effects of compression by viewing $\PhiT$ as inducing a form of compressed linear regression, where both the input data and targets are compressed.

\subsubsection{Preliminaries}

This approach is justified by noting that, as discussed in Section \ref{sec:learningpsrs}, in equations \eqref{eq:cinfdef} and \eqref{eq:Caodef} of our learning algorithm we are in fact performing implicit linear regression.
That is, for $(\UU,\Sing,\VV) = \textrm{SVD}(\cPthe)$:
\begin{align}\label{eq:isregress}
\VV\Sing^{-1} &= (\UU^\top\cPthe)^\dagger\\
		&= (\UU^\top\PhiT\Pthe\PhiH^\top)^\dagger.
\end{align}
In other words, $\VV\Sing^{-1}$ is the Moore-Penrose pseudoinverse of $\UU^\top\PhiT\Phe\PhiH^\top$, and multiplication by $\VV\Sing^{-1}$ is thus equivalent to performing least-squares linear regression.

Following the discussion in the previous section and to avoid unnecessary complication, we assume $\PhiH$ has orthonormal columns (i.e., is not compressive) while analysing the effects of compressing the tests.
In the case where $\PhiH$ has orthonormal columns, we define
$\boldsymbol{\Sigma}_{\TT,ao,\HH}$ as the compressed analogue of $\Pthao$, and see that \eqref{eq:Caodef} can be rewritten as 
\begin{align}
\Cao &= (\UU^\top\hat{\boldsymbol{\Sigma}}_{\TT,ao,\HH})(\UU^\top\cPthe)^\dagger\label{eq:startphidis}\\
	&= (\UU^\top\PhiT\Pthaoe\PhiH^\top)(\UU^\top\PhiT\Pthe\PhiH^\top)^\dagger \\ 
	&=(\UU^\top\PhiT\Pthaoe)\PhiH^\top(\PhiH^\top)^\dagger(\UU^\top\PhiT\Pthe)^\dagger \label{eq:phitimesphi}\\
	&=(\UU^\top\PhiT\Pthaoe)(\UU^\top\PhiT\Pthe)^\dagger\label{eq:endphidis},
\end{align}
where \eqref{eq:phitimesphi}-\eqref{eq:endphidis} holds since $\PhiH$ is assumed to have orthonormal columns.
An analogous result holds for $\cinf$ and thus, $\PhiH$ can, indeed, be omitted in our analysis (under the assumption that $\PhiH^\top\PhiH = \mb{I}$).

Moreover, we ignore the $\UU^\top$ term in what follows, which is justified in the case where $d' = d_{\TT}$ (i.e., when the truncated SVD dimension is equal to the test compression dimension). 
This $d' = d_{\TT}$ condition is very mild in the sense that the use of SVD during learning is primarily motivated by the need to efficiently compute pseudoinverses, which facilitates the efficient batch and incremental model-learning algorithms.
That is, the SVD is not used as a dimensionality reduction technique, as random projections are used in that role.\footnote{As noted in Section \ref{sec:stability} it is sometimes beneficial to use $d' < d_{\TT}$ and/or discard very small singular values in order to improve the numerical stability of computing inverses during learning. However, this issue of numerical stability is orthogonal to the analysis presented in this section.}   
Thus, under the assumption that $d' = d_{\TT}$, we have that
\begin{equation}
\mat{Ax} = \mat{b} \Rightarrow  \UU^\top\mat{A}\mat{x} = \UU^\top\mat{b}
\end{equation}
holds, since $\UU^\top$ is orthonormal  for  $d' = d_{\TT}$.
Thus, the appearance of $\UU$ in the pseudoinverse is inconsequential in an analysis of the effect of compressing prior to regression.

To simplify the analysis one step further, we assume that our test set is a minimal core set $\QQ$. Therefore, random projections are applied on $\ePqh$ and $\ePqoh$ matrices. The projections from over-complete test sets with rank bigger than $|\QQ|$ down to $d_\TT$ dimensions can be achieved by first projecting to size $|\QQ|$ and then projecting from $|\QQ|$ to $d_\TT$. 
By the results of Section \ref{sec:consistency}, this first projection leads to a consistent model, i.e. a model that is a linear transform of the model learned directly from $\ePqh$ and $\ePqoh$ matrices, since $\UU^\top\PhiT\mb{R}$ is invertible with probability 1 when the projected dimension is equal to $|\QQ|$ \citep{Boots:2009}.
The assumption that we work with the $\ePqh$ and $\ePqoh$ matrices directly (as apposed to invertible transforms of them) simplifies the analysis below in that we can elucidate our sparsity assumptions etc. directly in terms of the minimal core set of tests instead of random linear functions of tests in the minimal core set.   
This assumption is mild in that we could work with these random invertible linear transforms and discuss the discrepancy between a ``random'' TPSR (i.e., a TPSR defined via a random linear transform) and a compressed version of this ``random" TPSR, and this discussion would be analogous to that which is provided below, albeit with more cumbersome and unnecessarily complex derivations.
The assumption that we work with the minimal core set of tests simply allows for a more interpretable and less cluttered analysis.

Now, we define:
\begin{align}
\Bao = \Pqoh (\Pqh)^\dagger,\,\,\,\, \binf = (\Pqh)^\dagger \Phe.
\end{align}
Since $\QQ$ is a minimal core set of tests, the above is a TPSR representation~\citep{Boots:2009,Rosencrantz:2004}. Assume we have enough histories in $\HH$ such that matrices are full rank. Defining $\Pqhh$ and $\Pqohh$ to be the vectors containing the joint probabilities of all tests in the minimal core set and a fixed history $h$, we have (by the linearity of PSRs):
\begin{align}
\forall h: \Pqohh = \Bao \Pqhh,\,\,\,\, \Phh = \binf^\top \Pqhh.
\end{align}
One can thus think of finding the $\Bao$ and $\binf$ parameters as regression problems, having the estimates of $\Pqhh$s as noisy input features. We also have noisy observations of the outputs $\Pqohh$ and $\Phh$. Since the sample set suffers from the error in variables problem (i.e.,  is noisy both on the input and output values) direct regression in the original space might result in large estimation error. Therefore, we apply random projections, reducing the estimation error (variance) at the cost of a controlled approximation error (bias). And we get the added benefit that working in the compressed space also helps with the computation complexity of the algorithm.

Note that there is an inherent difference between our work and the TPSR framework. In TPSR, one seeks to find concise linear transformations of the observation matrices, whereas CPSR seeks to find good approximations in a compressed space (which cannot be linearly transformed to the original model). That said, approximate variants of the TPSR learning algorithm have been analyzed from the perspective of compressed regression (albeit without appealing to the compressed sensing framework we employ) \citep{kulesza2014low, boots2010predictive}. For example, \cite{kulesza2014low} analyze low-rank TPSR models where the rank of the learned model is made less than $|\QQ|$ by removing the least significant singular vectors of $\Pth$. 
We reiterate, however, that these analyses are distinct from the analysis presented in this work, as we analyze low-rank models where the rank is reduced via random projection-based compression (not by removing least-significant singular vectors). The following sections provide an analysis of the error induced by this compression and how the error propagates through the application of several compressed operators.

\subsubsection{Error of One Step Regression}

When the size of the projections is smaller than the size of the minimal core set, we have the implicit regression performed on a compressed representation. The update operators are thus the result of compressed ordinary least-squares regression (COLS). There are several bounds on the excess risk of regression in compressed spaces \citep{Maillard:2009, Maillard:2012, Fard:2012, Fard:2013}. In this section, we assume the existence of a generic upper bound for the error of COLS.

Assume we have a target function $f(\mb x) = \mb x^\top \mb w + b(\mb x)$ where $\mb x$ is in a $k$-sparse $D$-dimensional space, and $b(\cdot)$ is the bias of the linear fit. We observe an i.i.d. sample set $\{ (\mb x_i, f(\mb x_i)+\eta_i) \}^n_{i=1}$, where $\eta_i$'s are independent zero-mean noise terms for which the maximum variance is bounded by $\sigma^2_\eta$, and $\mb x_i$'s are sampled from a distribution $\rho$. Let $\hat f_d(\mb x)$ be the compressed least-squares solution on this sample with a random projection of size $d$.
That is, $\hat f_d(\mb x) = \mb{x}^\top\mb{\Phi}^\top \hat{\mb w}_d$ with 
\begin{equation}
\hat{\mb{w}}_d = (\mb\Phi \mb{X}^\top\mb{X}\mb\Phi^\top)^{-1}(\mb\Phi\mb{X}^\top)\mb{y} \in \R^{d},
\end{equation} 
where $\mb{X} \in \R^{n \times D}$ is a design matrix, $\mb y \in \R^n$ is a vector of training targets, and $\mb{\Phi} \in \R^{d \times D}$ is a random projection matrix.
 Define $\|g(\mb x)\|_{\rho(\mb x)} = \sqrt{\mathbb{E}_{\mb x \sim \rho}(g(\mb x))^2}$ to be the weighted $L^2$ norm under the sampling distribution. We assume the existence of a generic upper bound function $\epsilon$, such that with probability no less than $1-\delta$:
\begin{align}
\hspace{-0.5em}\|f(\mb x) - \hat f_d(\mb x)\|_{\rho(\mb x)} \leq \epsilon(n, D, d, \|\mb w\|^2, \|\mb x\|^2_{\rho(\mb x)}, \|b(\mb x)\|^2_{\rho(\mb x)}, \sigma^2_\eta, \delta). \label{eqn:colsbound}
\end{align}

The effectiveness of the compressed regression is largely dependent on how the $\|\mb w\| \|\mb x\|_{\rho(\mb x)}$ term behaves compared to the norm of the target values. We refer the reader to the discussions in \citet{Maillard:2009} and \citet{Maillard:2012} on the $\|\mb w\| \|\mb x\|_{\rho(\mb x)}$ term. In the case of working with PSRs, we have that the probability of the tests are often highly correlated. Using this property, we will show that $\|\mb w\|^2$ can be bounded well below its size.

In order to use these bounds, we need to consider the sparsity assumptions in our compressed PSR framework. We formalize the inherent sparsity, discussed in previous sections, as follows: For all $h$, $\Pqhh$ and $\Pqohh$ are $k$-sparse. Given that the empirical estimates of zero elements in these vectors are not noisy, for $\Delta_x = \ePqhh - \Pqhh$ we have that $\Delta_x$ is $k$-sparse (with a similar argument for $\Delta_y = \ePqohh - \Pqohh$).

%Finally, we assume that for all observations, $\Bao \Delta_x$ and $\Bao \Delta_y$ are $k'$-sparse. 

%These assumptions are justified in many domains. For instance, in the PocMan domain described below, we found the average column sparsity of all matrices to be at most 0.018\% (i.e. approximately 0.018\% of entries were non-zero).

To simplify the analysis, in this section we define our $\Cao$ matrices to be slightly different from the ones used in the described algorithm. By forcing the diagonal entries to be 0, we avoid using the $i$th feature for the $i$th regression. This removes any dependence between the projection and the target weights and simplifies the discussion. 
Since we are working with random compressed features as input, all of the features have similar correlation with the output, and thus removing one of them changes the error of the regression by a factor of $O(1/d)$. We can nevertheless change the algorithm to use this modified version of the regression so that the analysis stays sound.

%The resulting error bound should only change slightly if we use the original definitions.

The following theorem bounds the error of a one step update using the compressed operators. We use i.i.d. normal random projection for simplicity. The error bounds for other types of random projections should be similar.\footnote{The core modifications necessary are analogous to those used made in \cite{Achlioptas:2001} to adapt the Johnson-Lindenstrauss lemma to more general random matrices.} 
Let $[\mb{A}]_{-i,*}$ be matrix $\mb A$ with the $i$th row removed. We have the following:
\begin{theorem} \label{theorem:cao-error}
Let $\HH$ be a large collection of sampled histories according to $\rho$, and let $\mb{\Phi}^{d \times |\QQ|}$ be an i.i.d. normal random projection: $\mb{\Phi}_{ij} \sim \mathcal{N}(0,1/d)$. We observe noisy estimate $\ePqhh = \Pqhh + \Delta_x$ of input and $\ePqohh = \Pqohh + \Delta_y$ of the output, where elements of $\Delta_x$ and $\Delta_y$ are independent zero-mean random variables with maximum variance $\sigma^2_x$ and $\sigma^2_y$ respectively. Let $\sigma^2_1 \dots \sigma^2_{|\QQ|}$ be the decreasing eigenvalues of $E_{\rho(h)}[\Pqohh \Pqohh^\top]$. Choose $1 \leq m \leq |\QQ|$ such that $\sigma^2_m \leq 1$ and define $\nu = \sum^{|\QQ|}_{i=m+1}\sigma^2_i$. For $1\leq i \leq d$, define:
\begin{align}
\mb u_i = \mb{\Phi}_i \ePqoh (\mb{\Phi}_{-i}\ePqh )^{\dagger}.  \nonumber
\end{align}
Define $\Cao$ to be a $d\times d$ matrix such that:
\begin{align}
(\Cao)_i = [\mb u_{i,1}, \mb u_{i,2}, \dots, \mb u_{i,i-1}, 0, \mb u_{i,i}, \mb u_{i,i+1}, \dots, \mb u_{i,d-1}] .  \nonumber
\end{align}
Then with probability no less than $1-\delta$ we have:
\begin{align}
\left\| \Cao (\mb{\Phi}\Pqhh)  - \mb{\Phi} \Pqohh \right\|_{\rho(h)}  
\leq \sqrt d \epsilon(|\HH|, |\QQ|, d, w^2, x^2, b^2, \sigma_\eta^2, \delta/4d),
\end{align}
where:
\begin{eqnarray}
w^2 &=& \|\Bao\|^2(m + 4 \sqrt{m} \ln (4d/\delta)),\\
x^2 &=& \|\Pqhh\|^2_{\rho(h)},\\
b^2 &=& \nu + 4 \sqrt \nu \ln(4d/\delta),\\
\sigma_\eta^2 &=& \frac{4 k \ln(4|\QQ|/\delta)}{d} \sigma^2_y + w^2 \sigma^2_x.
\end{eqnarray}
\end{theorem}

The proof is included in the appendix. The main idea of the theorem is to use the dependence and sparsity of the features to tighten the bound on the error of compressed regression. When most of the variation in the PSR state can be explained using $m$ linear observations, we can substitute the $\mb{\Phi}_i \Bao$ target weights having norm $O(\sqrt{|\QQ|})$, with a linear approximation having much smaller norm $O(\sqrt m)$, at the expense of a small bias $b$. The theorem also describes the overall noise combining the effects of $\Delta_x$ and $\Delta_y$.

Theorem~\ref{theorem:cao-error} has three main implications. One is that the complexity of the compressed regression depends on how fast the eigenvalues drop for the minimal core set covariance matrix. If the eigenvalues drop exponentially fast, as is observed empirically in our experiments, we can guarantee a smaller regression error. Second, if the projection size is of order $O(k \ln |\QQ|)$ we can control the variance of the combined noise term. Third, if we use the sparse COLS bound of \citet{Fard:2012,Fard:2013}, we can can show that regression of size $O(k \ln |\QQ|)$ should be enough to decrease the overall estimation error at the expense of a controlled bias.

The following corollary follows immediately from Theorem~\ref{theorem:cao-error} by union bounding over all action-observation pairs.
\begin{corollary} \label{corollary:cao-error}
Using the assumptions of Theorem~\ref{theorem:cao-error}, with probability no less than $1-\delta$ we have for all $a \in \mathcal{A}$ and $o \in \mathcal{O}$:
\begin{align}
\left\| \Cao (\mb{\Phi}\Pqhh)  - \mb{\Phi} \Pqohh \right\|_{\rho(h)}  
\leq \sqrt d \epsilon(|\HH|, |\QQ|, d, w^2, x^2, b^2, \sigma_\eta^2, \delta/(4d|\mathcal{A}| |\mathcal{O}|)),
\end{align}
where $w^2 = \max_{ao}\|\Bao\|^2(m + 4 \sqrt{m} \ln (4d/\delta))$, and other factors are as defined in Theorem~\ref{theorem:cao-error}.
\end{corollary}

\subsubsection{Error of the compressed normalizer}
The $\cinf$ operator is the normalization operator for the compressed space. Therefore, for any history $h$, $\cinf^T \mb{\Phi} \Pqhh$ should equal $\Phh$. The following theorem provides a bound over the error of such a prediction:

\begin{theorem} \label{theorem:cinf-error}
Let $\HH$ be a large collection of sampled histories according to $\rho$. We observe noisy estimate $\ePqh = \Pqh + \Delta_x$ of input and $\Ph = \Phe + \Delta_z$ of the output, where elements of $\Delta_x$ and $\Delta_z$ are independent zero-mean random variables with maximum variance $\sigma^2_x$ and $\sigma^2_z$ respectively. Define $\cinf = (\mb{\Phi}_i \ePqh)^\dagger \Ph$. Then with probability no less than $1-\delta$ we have:
\begin{align}
&\left\| \cinf^\top (\mb{\Phi}\Pqhh)  - \Phh \right\|_{\rho(h)} \leq \epsilon(|\HH|, |\QQ|, d, \|\binf\|^2, \|\Pqhh\|^2_{\rho(\mb h)}, 0, \sigma_\infty^2, \delta), \nonumber
\end{align}
where we define effective noise variance $\sigma_\infty^2 = \sigma^2_z + \sigma^2_x \|\binf\|^2$.
\end{theorem}

The proof is included in the appendix.

\subsubsection{Error Propagation}
Once we have the one step errors of compressed operators, we can analyse the propagation of errors as we concatenate the operators. Define $o_{1:n} = o_1 o_2 \dots o_n$ (and similarly for $a_{1:n}$ and $[ao]_{1:n}$). We would like to bound the error between $\PP(o_{1:n} ||  a_{1:n})$ and our prediction $\cinf \mb C_{a_{n}o_{n}} \mb C_{a_{n-1}o_{n-1}} \dots \mb C_{a_{1}o_{1}} \mb c_1$.

\newcommand{\Pqaot}{\boldsymbol{\mathcal{P}}_{\QQ,[ao]_{1:t}}}
\newcommand{\Pqaotm}{\boldsymbol{\mathcal{P}}_{\QQ,[ao]_{1:t-1}}}

\newcommand{\Pqaon}{\boldsymbol{\mathcal{P}}_{\QQ,[ao]_{1:n}}}
\newcommand{\Pqaonm}{\boldsymbol{\mathcal{P}}_{\QQ,[ao]_{1:n-1}}}

Since the theorems in the previous sections were in terms of a fixed measure $\rho$, we have to make distributional assumptions to simplify the derivations. Assume that we fit our model using samples $h \sim \rho$, imposing a distribution $\Pqhh \sim \mu$. Note that as we increase the size of a history $h$, the norm of $\Pqhh$ becomes smaller. We make the assumption that for all $1 \leq t \leq n$, for a history $[ao]_{1:t} \sim \rho_t$, the implied $\Pqaot$ is sampled from a scaled version of $\mu$ (i.e., $\frac{1}{s_t}\Pqaot \sim \mu$). Therefore $\|f(\Pqhh)\|_{\rho_t(h)} = \|f(s_t\Pqhh)\|_{\rho(h)}$.

\begin{theorem} \label{theorem:prop-error}
Let $\epsilon$ and $\epsilon_\infty$ be the bounds of Corollary~\ref{corollary:cao-error} and Theorem~\ref{theorem:cinf-error} respectively, for a sample $\HH$ according to $\rho$ and failure probability $\delta/2$. Let $\rho_n$ and its marginals $\rho_{n-1} \dots \rho_1$, be distributions over histories of size $n, n-1, \dots 1$ respectively, such that $\|f(\Pqhh)\|_{\rho_t(h)} = \|f(s_t\Pqhh)\|_{\rho(h)}$ for all measurable $f$. With probability $1-\delta$:
\begin{equation}
\left\| \cinf \mb C_{a_{n}o_{n}} \mb C_{a_{n-1}o_{n-1}} \dots \mb C_{a_{1}o_{1}} \mb c_1 - \PP(o_{1:n} || a_{1:n}) \right\|_{\rho_n} \; \leq \; \epsilon_\infty s_n + \|\cinf\| \epsilon \sum^{n-1}_{t=1} s_t c^{n-t-1},
\end{equation}
where $c = \max_{a,o} \|\mb C_{ao}\|$.
\end{theorem}

The proof is included in the appendix. Note that $s_t$ is exponentially decreasing in $t$ (because longer tests are less probable). The norm of the update operators are expected to be less than 1 (as they shrink the vector of test probabilities). Combining these two, we expect the summation in the bound of Theorem~\ref{theorem:prop-error} to be over a small exponential function of $n$.

\section{Planning with CPSRs}\label{sec:planning}

The learning algorithm presented in Section \ref{sec:learningcpsrs} facilitates the construction of accurate predictive models in large complex partially observable domains.
In this section, we outline how to plan (near)-optimal sequences of actions using such a learned model. 
The planning approach we employ was first proposed by \cite{Ong:2012}.
In essence, the approach substitutes a predictive state in place of an observable state in the standard fitted-$Q$ learning algorithm of \cite{Ernst:2005}. 

Unlike point-based value-iteration PSR (PBVI-PSR) planning algorithms, the theoretical convergence of the fitted-$Q$ algorithm does not require that the PSR correspond to a finite-dimensional POMDP. 
That is, existing error-bounds for PBVI-PSR require that the PSRs used in planning correspond to some finite-dimensional POMDP \citep{Izadi:2008}, whereas in general PSRs may have no corresponding finite-dimensional POMDP \citep{denis2008rational}.\footnote{It is worth noting, however, that the PSR-PBVI error bounds could possibly be modified to alleviate this issue and that PBVI-PSR algorithms have been employed with considerable empirical success \citep{Boots:2009, Izadi:2008}.}
In contrast, the fitted-$Q$ approach only requires that the input state-space be sufficient to describe the system, and PSRs satisfy this requirement, meaning that the convergence results for fitted-$Q$ carry over to the PSR setting (when an exact PSR model is used) \citep{Ernst:2005}.\footnote{The error bounds for PSR-PBVI also require that an exact model is known. In general, current theoretical results on PSR planning ignore the impact of estimation and/or approximation errors incurred during model-learning, though empirical analyses (e.g., the work of \cite{Boots:2009} and Section \ref{sec:empirical} of this paper) suggest that the impact of such errors is small.}
Moreover, the fitted-$Q$ approach does not explicitly require learning a model of rewards prior to the application of the planning algorithm (i.e., the reward model is captured only through the $Q$-function).
We found this to be preferable to explicitly modelling the immediate rewards as a function of the CPSR states prior to planning, as such an explicit model introduces an extra (and unnecessary) level of approximation.
In what follows, we briefly review the fitted-$Q$ approach and provide a high-level description of our planning algorithm.

\subsection{Fitted-$Q$ with CPSRs}  

\begin{algorithm}{Algorithm 1: Fitted-$Q$ with CPSR}
\begin{algorithmic}
\STATE \textbf{Inputs:} A set $\mathcal{D}$ of tuples of the form $(\mb{c}_t,a_t,r_t,\mb{c}_{t+1})$ constructed using a CPSR model, where $r_t$ is a numerical reward; $\mathcal{R}$, a regression algorithm; $\gamma$, a discount factor; and $T$, a stopping condition
\STATE \textbf{Outputs:} A policy $\pi$
\end{algorithmic}
\vspace{1pt}
\begin{algorithmic}[1]
\STATE{$k \leftarrow 0$}
\STATE{Set $\hat{Q}_k(\mb{c}_t,a) = 0 \:\: \forall a \in \Acts$ and all possible $\mb{c}_t$}
\REPEAT
\STATE{$k \leftarrow k + 1$}
\STATE{Build training set, $\mathbb{T} = \{(y^l,i^l), l=1,...,|\mathcal{D}|\}$ where:
$i^l = (\mb{c}^l_t, a^l_t)$ and $y^l = r^l_t + \gamma\max_a\hat{Q}_{k-1}(\mb{c}_{t+1}^l,a)$} 
\STATE{Apply $\mathcal{R}$ to approximate $\hat{Q}_k$ from $\mathbb{T}$}
\UNTIL{$T$ is met}
\OUTPUT{$\pi$, where $\pi(\mb{c}_t) = \textrm{argmax}_a\{\hat{Q}_k(\mb{c}_t,a)\}$}
\end{algorithmic}
\end{algorithm}

As stated above, fitted-$Q$ with PSRs is analogous to the MDP case, with the predictive state taking the place of the MDP state in Algorithm 1.
The algorithm iteratively builds more and more accurate approximations of the $Q$-function, which in our case maps predictive states and actions to expected returns. 
In this work, the \textit{Extra-Trees} algorithm is used as the base regression algorithm \citep{Geurts:2006}, as it is a non-linear function approximator for which the fitted-$Q$ convergence results hold \citep{Ernst:2005}.
For $T$, the termination condition, we use an iteration limit (instead of an $\epsilon$ convergence condition), as this allows for more accurate predictions of runtimes.

Letting $\Psi(T)$ be the expected number of iterations under stopping condition $T$ and assuming that the splitting procedure for nodes in the Extra-Trees algorithm takes constant time, the computational complexity of this fitted-$Q$ approach is (recalling the definitions of Section \ref{sec:learningcpsrs}): 
\begin{equation}\label{eq:fittedqruntime}
O\left(\Psi(T)\times L|Z|log\left(L|Z|\right)\right),
\end{equation}
which is a factor $\Psi(T)\times log(L|Z|)$ greater than the complexity for the model-learning algorithm of Section \ref{sec:learningcpsrs}.
In practice, we found Algorithm 1 to be several orders of magnitude slower than the CPSR learning algorithm.

\subsection{Combined Learning and Planning}

Algorithm 2 specifies how CPSR model-learning and the fitted-$Q$ planning algorithm are combined at a high level.
This general specification permits a variety of sampling and $Q$-function approximation strategies.
Specifically, it permits pure unbiased random sampling, interleaving exploration and exploitation phases, or even the drawing of samples from some arbitrary (e.g., expert) policy.
Of course, if non-blind policies are used then the sample distribution becomes biased (i.e., the samples are no longer i.i.d.), and the analysis of Section \ref{sec:theory} no longer holds.

%The specification permits using either an unbiased random exploration approach (when $N=0$) or a goal-directed sampling approach to gather the trajectories necessary to learn the CPSR. 
%The benefits of the goal-directed approach are elucidated by \cite{Ong:2012}.
%Intuitively, the approach is more sample efficient as it restricts the learning to areas of the state space relevant to achieving goals.
%In this work this sample efficiency is less important in terms of model learning, however, since the efficient CPSR learning approach is linear in the number of samples.
%Nonetheless, the fitted-$Q$ algorithm is still reasonably expensive, so the goal-directed sampling may provide benefits in some settings.

Also note that the number of iterations used by the learner and planner need not be identical. 
More specifically, more samples may be used to learn the CPSR model than are used in planning.
This is a pragmatic specification, as the CPSR learning algorithm can efficiently accommodate orders of magnitude larger sample sets than the fitted-$Q$ planner (by \eqref{eq:cpsrcomplex} and \eqref{eq:fittedqruntime}).

\begin{algorithm}{Algorithm 2: Combined learning and planning}
\begin{algorithmic}
\STATE \textbf{Inputs:} $\pi_s$, a sampling policy; $N$, the number of sampling iterations; $I_m$, the number of trajectories to use in learning; and $I_p$, the number of trajectories to use in planning ($I_m \geq I_p$)
\STATE \textbf{Outputs:} A CPSR model, $\mb{C}$ and policy $\pi$
\end{algorithmic}
\vspace{1pt}
\begin{algorithmic}[1]
\STATE{$\mathcal{D}_0 \leftarrow \emptyset$} 
\STATE{Initialize the CPSR model, $\mb{C}$}
\FOR{i=1 to N}
\STATE{Sample $I_m$ trajectories, $Z_{i}$, using $\pi_s$}
\STATE{Update $\mb{C}$ using $Z_{i}$}
\STATE{Sub-sample $I_p$ trajectories from $Z_i$ and use $\mb{C}$ to construct a tuple-set $\mathcal{D}_{i}$}
\STATE{$\mathcal{D}_{i} \leftarrow \mathcal{D}_{i} \cup \mathcal{D}_{i-1}$}
\STATE{Apply Algorithm 1 with $\mathcal{D}_{i}$ to learn a policy, $\pi_{i}$}
\STATE{[Optional] Update $\pi_s$ (e.g., using $\pi_{i}$)}
\ENDFOR
\OUTPUT{$\mb{C}$ and $\pi_{N}$}
\end{algorithmic}
\end{algorithm}

\section{Empirical Results}\label{sec:empirical}

We examine empirical results pertaining to both the model quality of compressed models and the proficiency of model-based planning.
The goal of this analysis in the model-quality setting is to elucidate (1) the empirical cost (in terms of prediction accuracy) of performing compression (if any), (2) the compute-time reduction engendered by the use of compression, and (3) the impact of the implicit regularization induced by performing compression.
We also provide model-quality results explicitly comparing prediction performance when histories are compressed versus uncompressed, showing that history compression has a negligible effect empirically (and justifying the simplifying assumption that $\PhiH^\top\PhiH = \mb{I}$ in Section \ref{sec:theory}).

In the planning setting, we again seek to elucidate the empirical impact of performing compression, and we do so using three different partially observable domains.
First, we use a simple synthetic robot navigation domain (identical to that used in the model-quality experiments) to compare the planning performance of agents trained with CPSR models, agents trained with uncompressed TPSR models, and memoryless (model-free) agents, which serve as a baseline.
Next, we examine a massive partially observable domain that is intractable for classic POMDP-based approaches, demonstrating how the use of compression facilitates learning and planning in settings where it would be otherwise intractable.  We also provide a qualitative comparison to the Monte-Carlo AIXI algorithm \citep{Veness:2011}, a related model-based reinforcement learning approach, using this domain.
Lastly, we apply CPSR based learning and planning to the difficult real-world task of adaptive migratory management \citep{Nicol:2013}. 
In this adaptive migratory management problem, a sequential decision-making agent must learn a model of how a certain bird species migrates and how their migration patterns are adversely affected by rising sea-levels (and must do so without prior domain-specific knowledge). Using this learned model the agent must determine an optimal policy for  protecting different locations along the birds' migratory route so as to minimize population decline \citep{martin2007optimal,Nicol:2013}. 
This difficult real-world domain, which builds upon hand-crafted simulators and ecological datasets \citep{iwamura2011spatial,Nicol:2013}, demonstrates both the benefits of compression (in that it is computationally intractable for uncompressed TPSR) and the stark benefits of model-based planning over memoryless (model-free) planning.

\subsection{Projection Matrices}

In this analysis, we examine three different classes of random projection matrices: spherical, Rademacher, and hashed.
The spherical projection matrices contain random Gaussian distributed entries and are identical to those used in \cite{Hamilton:2013}.
The Rademacher are a related class of random matrices where each entry is an independent Rademacher variable; these matrices also satisfy the JL lemma \citep{Baraniuk:2009} and can afford additional efficiencies with low level implementations that exploit the fact that only additions and subtractions are used in the matrix multiplications (this optimization is not used here) \citep{Achlioptas:2001}.
The hashed random projection matrices induce a feature-mapping analogous to random hashing;  each column of the random projection matrix has a 1 in a random position and the other entries are zero.
These random hashing matrices do not directly satisfy the JL lemma, but they have been shown to preserve certain kernel-functions and perform extremely well in practice \citep{Weinberger:2009,Shi:2009}.

\subsection{Domains}

The domains used are based upon previous work on planning with PSRs and on model-based reinforcement learning in large, complex partially observable domains.

\subsubsection{ColoredGridWorld}

\begin{figure}[h]
\centering
\includegraphics[scale=0.55]{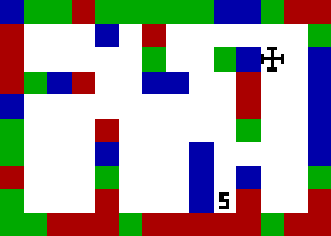}
\caption{Graphical depiction of \emph{ColoredGridWorld}. The \textbf{S} denotes the start position and the target denotes the goal.} 
\label{fig:modqual}
\end{figure}

The first domain, \emph{ColoredGridWorld}, is conceptually similar to the simulated robot navigation domains commonly used in the PSR literature and is a direct extension of the \emph{GridWorld} domain used in \cite{Hamilton:2013} and \cite{Ong:2012}.
The environment is a 47-state maze with coloured walls.
The agent must navigate from a fixed start state to a fixed goal state using only aliased local observations.
The action space consists of moves anywhere in the four cardinal directions (moving into walls produces no effect). To simulate noise in the agent's actuators, actions fail with probability $0.2$, and if this occurs, the agent moves randomly in a direction orthogonal to that which was specified. The observation space consists of whether or not the agent can see coloured walls in any of the 4 cardinal directions (one observation per wall).
There are three possible colors, so there are 3 possible observations per wall and thus $81$ possible observations in total.
A reward of $1$ is returned at the goal state (resetting the environment), and no other states emit rewards.

Though simple, this domain is quintessentially partially observable in that it is impossible to learn how to reach the goal without incorporating memory. 
Moreover, the added complication of coloured walls exponentially increases the cardinality of the observation space, leading to many possible tests and histories.
In essence, the agent cannot know a priori whether the colouring is pertinent to the problem, so it vastly complicates the learning problem.

\subsubsection{Partially Observable PacMan}

\begin{figure}[h]
\centering
\includegraphics[scale=0.45]{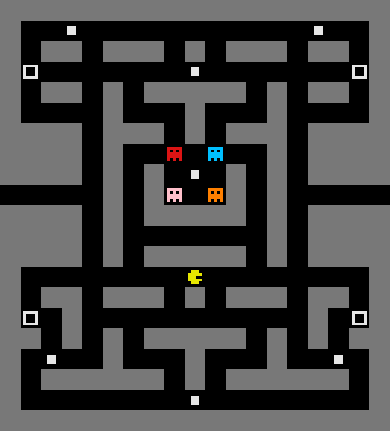}
\caption{Graphical depiction of \emph{S-PocMan}. The white dots denote food and the white annuli denote power-pills. The yellow PacMan figure denotes the fixed starting position} 
\label{fig:modqual}
\end{figure}

The second domain used is based upon the partially observable PacMan domain, denoted \textit{PocMan}, first proposed by \cite{Silver:2010}.
It is an extremely large partially observable domain with on the order of $10^{56}$ states \citep{Veness:2011}.
The basic dynamics follow that of the video-game PacMan: an agent must navigate a maze-like environment starting from a fixed start-point, collecting food and avoiding coming in contact with any of four ghosts.

In this work, we examine two versions of the domain. 
The first version is a replica of the \textit{PocMan} domain used by \cite{Veness:2011} in their work on a Monte Carlo AIXI approximation.
In the second version, which we call \textit{S-PocMan}, we further complicate the environment by dropping the parts of the observation vector that allow the agent to sense in what direction food lies, and we sparsify the amount of food in the environment.
In the original domain food was placed in each position with probability $\frac{1}{2}$; in \textit{S-PocMan} there are only 7 pieces of food in total, each in a fixed position.
The reason for examining this more difficult version of the domain is that, as summarized in Section \ref{sec:planningresults}, we found that a memoryless controller was able to perform extremely well on the original \emph{PocMan}, achieving results approaching that of the AIXI algorithm.
In other words, simply treating the original \emph{PocMan} domain as if it were fully observable led to very good results.
This seems to be due to the fact that the food rewards were plentiful and fully observable.
In \emph{S-PocMan} we make the problem more partially observable in order to demonstrate the usefulness of a model-based approach.

\subsubsection{Adaptive Migratory Management}

The last domain we examine is based upon the ecological task of \emph{adaptive migratory management} (\emph{AMM}).
The specific goal of \emph{AMM} is to use intervention to protect certain regions in a bird-species' migratory route.
In this work, we focus on the Lesser Sand Plover, which is one of many species that uses the East-Asian-Australasian (EAA) migratory route.
While migrating, the Lesser Sand Plover stop at staging sites where they feed on invertebrates and gather energy \citep{martin2007optimal}.
These staging sites are located at intertidal mudflats that are especially susceptible to rising sea levels \citep{iwamura2011spatial}.
The sites can be protected via intervention, but limited resources within the conservation community means that protection can only be implemented at a limited number of sites within a particular year.
By phrasing the task of protecting these intertidal areas as a sequential decision-making problem, the hope is to learn an optimal strategy for intervention.

In \cite{Nicol:2013} the \emph{AMM} problem is formalized, and a simulator based data-set is provided (for a number of species including the Lesser Sand Plover).
At its core the simulator uses a network-flow model for the migratory routes augmented with hand-crafted models for sea-level rises, population declines, and other relevant elements. See \cite{Nicol:2013} for a complete description.
In this work, we use data generated from the simulator, and we attempt to both learn a succinct model of the domain and optimize decisions using this learned model (i.e., we do not assume access to information contained within the internal simulator state).\footnote{Note that for the benchmark results presented in \cite{Nicol:2013}, they use knowledge of the underlying simulator state and cast the planning problem in the POMDP framework, while in this work we solve both the learning and planning problems (rather than just the planning problem).}

Formally, at each time point (which roughly corresponds to a year) the decision-making agent receives a vector of observations, where the first entry corresponds to the population level at the breeding site/node and the next three entries correspond to the protection levels at the three intertidal sites/nodes on the Lesser Sand Plover's migratory route. 
There are four discretized population levels and three protection levels, corresponding to protection against three increasing states of sea-level rise.
There are thus 108 unique possible observations.
At each time-step the decision-making agent must increase the protection level at one of the non-breeding nodes, and thus there are three possible actions at each time-step. (If the agent opts to protect a node which is already maximally protected then the action has no effect).
Internal dynamics of the underlying system-model determine how protection levels decline over time, but none of this information is available to the agent.
At the beginning of a simulation (i.e., in the fixed start-state) the protection and population levels are set to their minimal discretized values.

\subsection{Model Quality Results}

\begin{figure}[h]
\centering
\includegraphics[scale=0.65]{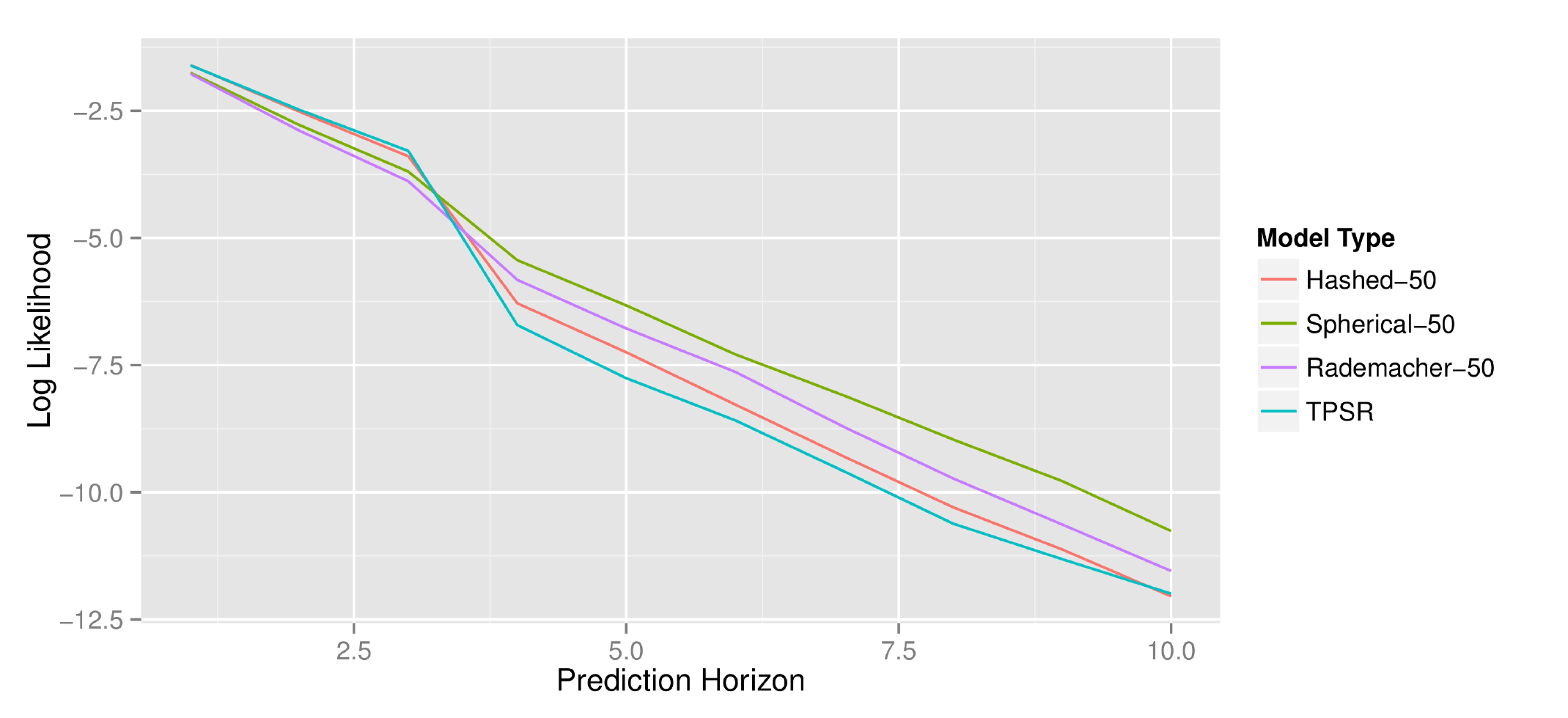}
\caption{Model-quality results on the \emph{ColoredGridWorld} domain. Plot shows the log-likelihood of the test data given the different models as the prediction horizon is increased. The numbers adjacent to the CPSR projection types correspond to the compressed dimension used. $95 \%$ confidence interval error bars are too small to be visible.} 
\label{fig:modqual}
\end{figure}

\begin{figure}[h]
\centering
\includegraphics[scale=0.47]{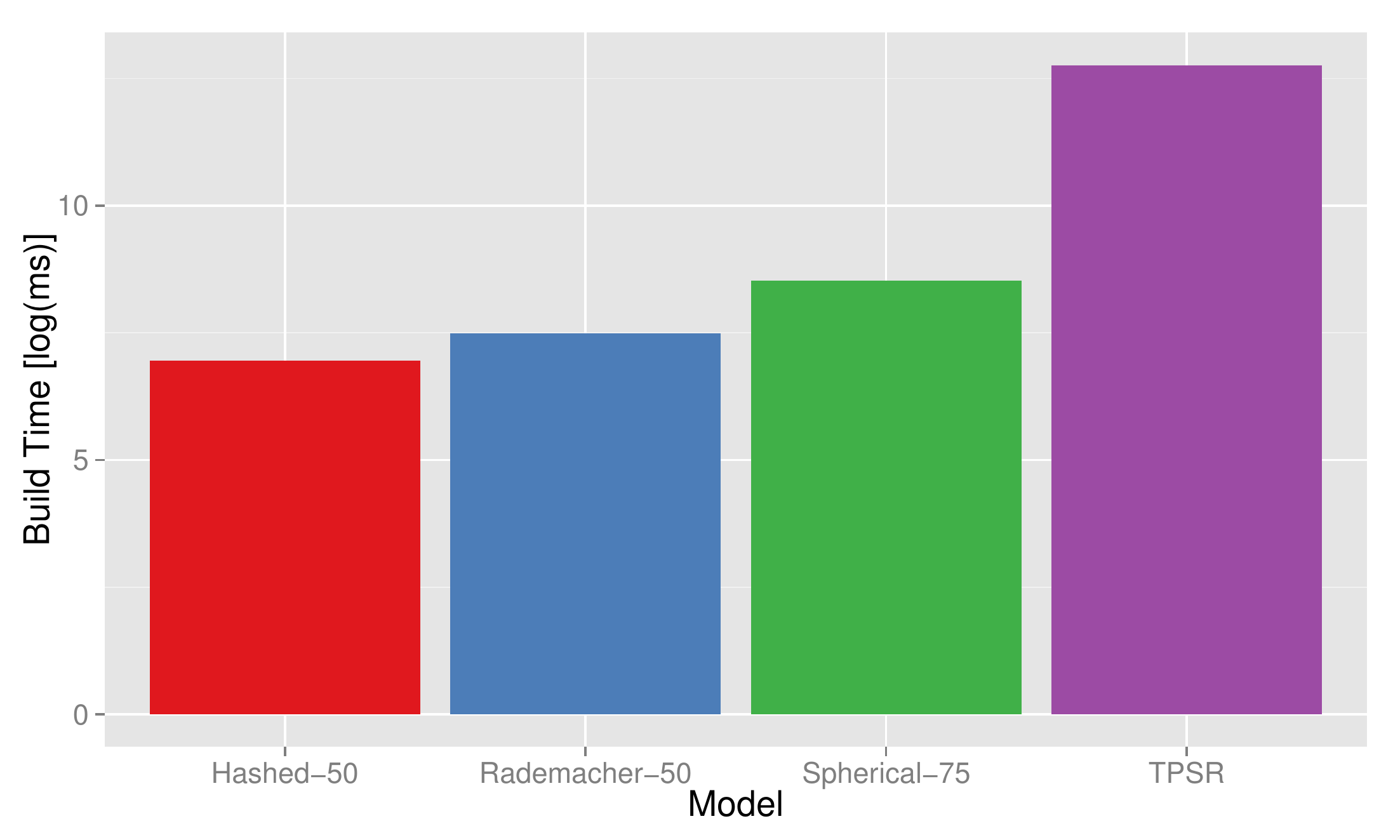}
\caption{Model build times (on a log-scale) for the different model types on the \emph{ColoredGridWorld} domain. Compressed dimension sizes are listed next to the model names. Times do not include time taken to build the training set. $95 \%$ confidence interval error bars are too small to be visible.} 
\label{fig:modbuildtimes}
\end{figure}

We examined the model quality of different CPSRs and an uncompressed TPSR on the \textit{ColoredGridWorld} domain.  
Sample trajectories were generated using a simple $\epsilon$-greedy exploration policy, where the non-random actions were determined by a policy learned via a memoryless controller.
All models were set with $d'=5$, where $d'$ is final model dimension (from Section  \ref{sec:learningcpsrs}) set after performing SVD; however, singular values below a tolerance of $10^{-6}$ were also discarded.
All tests, $\tau_i$, with $|\tau_i| \leq 7$ were included in the estimation process\footnote{If a particular test was never encountered in the training data, however, it was discarded, as such tests lead to singularities in the observable matrices.} (including longer length tests did not improve performance).
For the CPSR models,  we set $d_\TT = d_\HH$, as preliminary experiments did not reveal any significant benefits to using $d_\TT \neq d_\HH$ and examined projection dimensions in the range $[25,75]$. Only the best performing size (determined through cross-validation) is reported.
All models used $10000$ train trajectories (of max length 13) and were evaluated with $10000$ trajectories.
The PacMan-style domains and the \emph{AMM} domain were not examined in this model-quality context as naive TPSRs exhausted memory limits when tests of length longer than $1$ were used, making a rigorous comparison is infeasible.\footnote{Experiments were run on a machine with a 8-core 3.2 GHz Intel Xeon processor (x64 architecture) and 8Gb of RAM.}

Figure \ref{fig:modqual} plots the average log-likelihood of the models as the prediction horizon (i.e., length of the sequences to predict) is increased.
The log-likelihood for a single sequence is computed by taking the logarithm of the probability obtained via \eqref{eq:prediction}, and this likelihood is averaged over all sequences in the test set. 
From this figure, we see that the compressed models are not only competitive with the uncompressed TPSR, they actually outperform TPSR at longer prediction horizons.
We conjecture that this is due to the regularization induced by the use of random projections.
Figure \ref{fig:modbuildtimes} plots the build times for the different models, showing that the compressed models can be built in a fraction of the time required to build the uncompressed TPSR.

\begin{figure}[h!]
\centering
\includegraphics[scale=0.37]{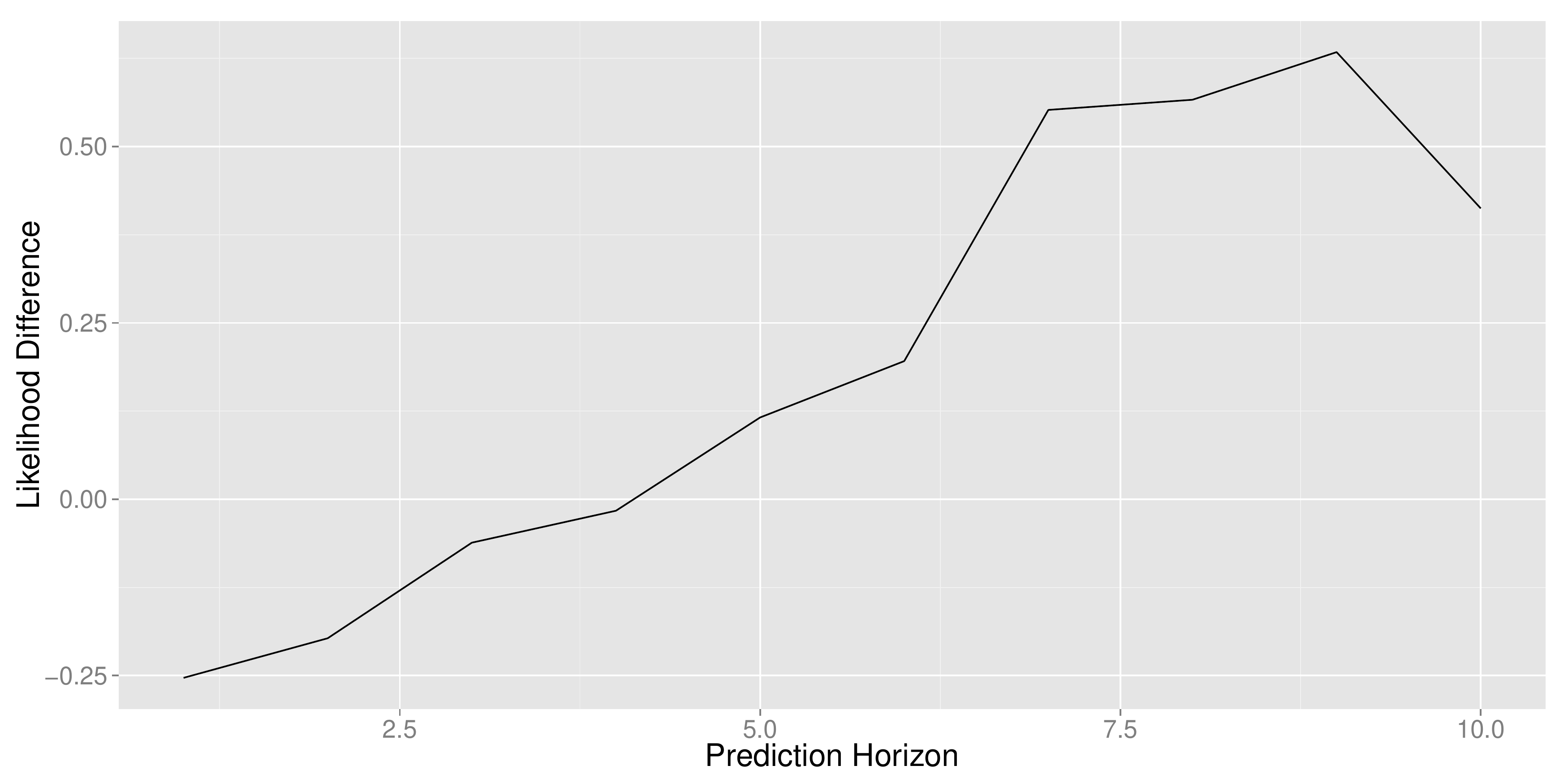}
\caption{Difference in log-likelihood between a model where histories are not compressed and a model where histories are compressed.}
\label{fig:histcompress}
\end{figure}

Figure \ref{fig:histcompress} shows a focused experiment examining the impact of compressing histories, compared to only compressing tests as was done in \cite{Hamilton:2013}. 
These results show $\log(\mathcal{L(\theta)}) - \log(\mathcal{L}(\theta_{HC}))$, the difference between the model-likelihood for a model where histories are not compressed ($\theta$) and where histories are compressed ($\theta_{HC}$).
Both the predictive models are constructed using spherical projection matrices and using (identical) samples generated from the \textit{ColoredGridWorld} domain (with the experimental set-up described above).
As is evidenced in the plot, there is only a small difference in likelihood between the two models (cf. the likelihood difference seen in figure \ref{fig:modqual}), and in fact, the model with compressed histories does slightly better for the first few time steps.

\subsection{Planning Results}\label{sec:planningresults}

\begin{figure}[h!]
\centering
\includegraphics[scale=0.4]{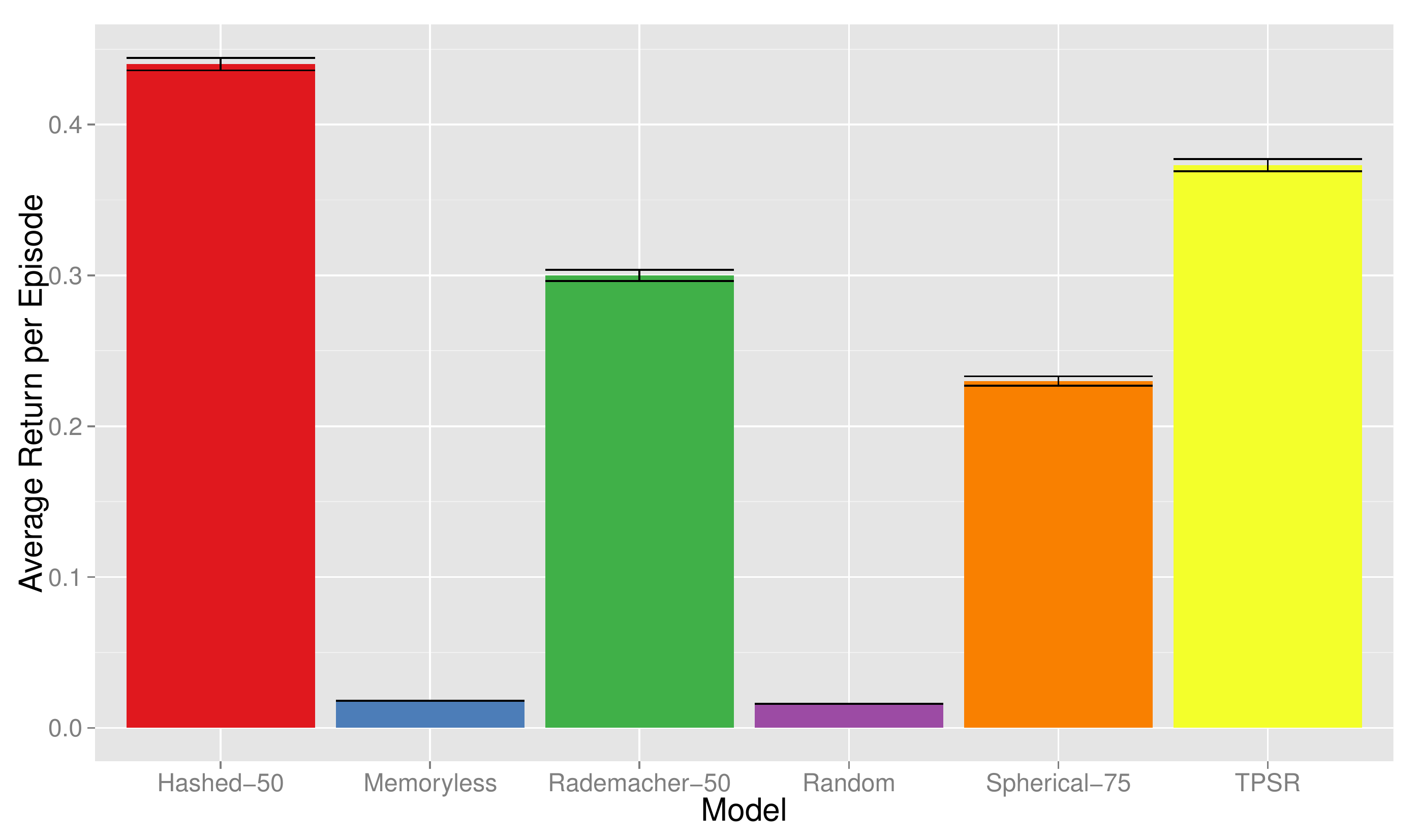}
\caption{Average return per episode achieved in the \textit{ColoredGridWorld} domain using different models and the baselines. Compressed dimension sizes are listed next to the model names. $95 \%$ confidence interval error bars are shown.} 
\label{fig:cgridworldplan}
\end{figure}

Next, we apply the full learning and planning approach (Algorithm 2) to the domains \textit{ColoredGridWorld, PocMan}, \textit{S-PocMan}, and \textit{AMM}.

In all experiments, we used 10000 random sampled trajectories to build the models and again used $d_\TT = d_\HH$.
For planning, we used $I_p=1000$ with $N=1$ and a random sampling strategy; this represents the standard unbiased batch-learning setting (Section \ref{sec:discuss} discusses the possibility of using more complex sampling strategies).  
And for the fitted-$Q$ algorithm, we used 100 fitted-$Q$ iterations, one \textit{Extra-Tree} ensemble of 25 trees per action, and the default settings for the \textit{Extra-Trees} \citep{Geurts:2006}.
As a baseline, we examined the performance of a memoryless controller on the domains.
This controller is analogous to treating the domains as fully observable and running the standard fitted-$Q$ algorithm of \cite{Ernst:2005}.
In order to achieve a fair comparison, the memoryless controller is permitted to use samples that would otherwise be used for model-learning in order to refine its policy (i.e., the memoryless baseline uses the same total number of samples in the experiments as the model-based methods).
The use of this baseline is not arbitrary, as its success provides an empirical measure of how partially observable a domain is with respect to planning; if a domain is easily solved by the memoryless controller then it is nearly fully observable in that immediate observations are sufficient for determining near-optimal plans.  
We also used a simple random planner which selects actions uniformly randomly as a second baseline.

\subsubsection{ColoredGridWorld}

For \textit{ColoredGridWorld}, the models examined were identical to those described in the model quality experiments above.
A discount factor of $\gamma=0.99$ was used for this domain.

Figure \ref{fig:cgridworldplan} details the performance of the different algorithms on the \textit{ColoredGridWorld} domain.
For this domain, the hashed CPSR algorithm achieved the best performance while the TPSR algorithm performed second-best.
All the PSR-based approaches vastly outperformed the memoryless-controller baseline.
This is expected, as the \textit{ColoredGridWorld} problem is strongly partially observable.

\subsubsection{Partially Observable PacMan}

\begin{figure}[h!]
        \centering
        \begin{subfigure}[b]{0.5\textwidth}
                \includegraphics[height=7cm,width=7.5cm]{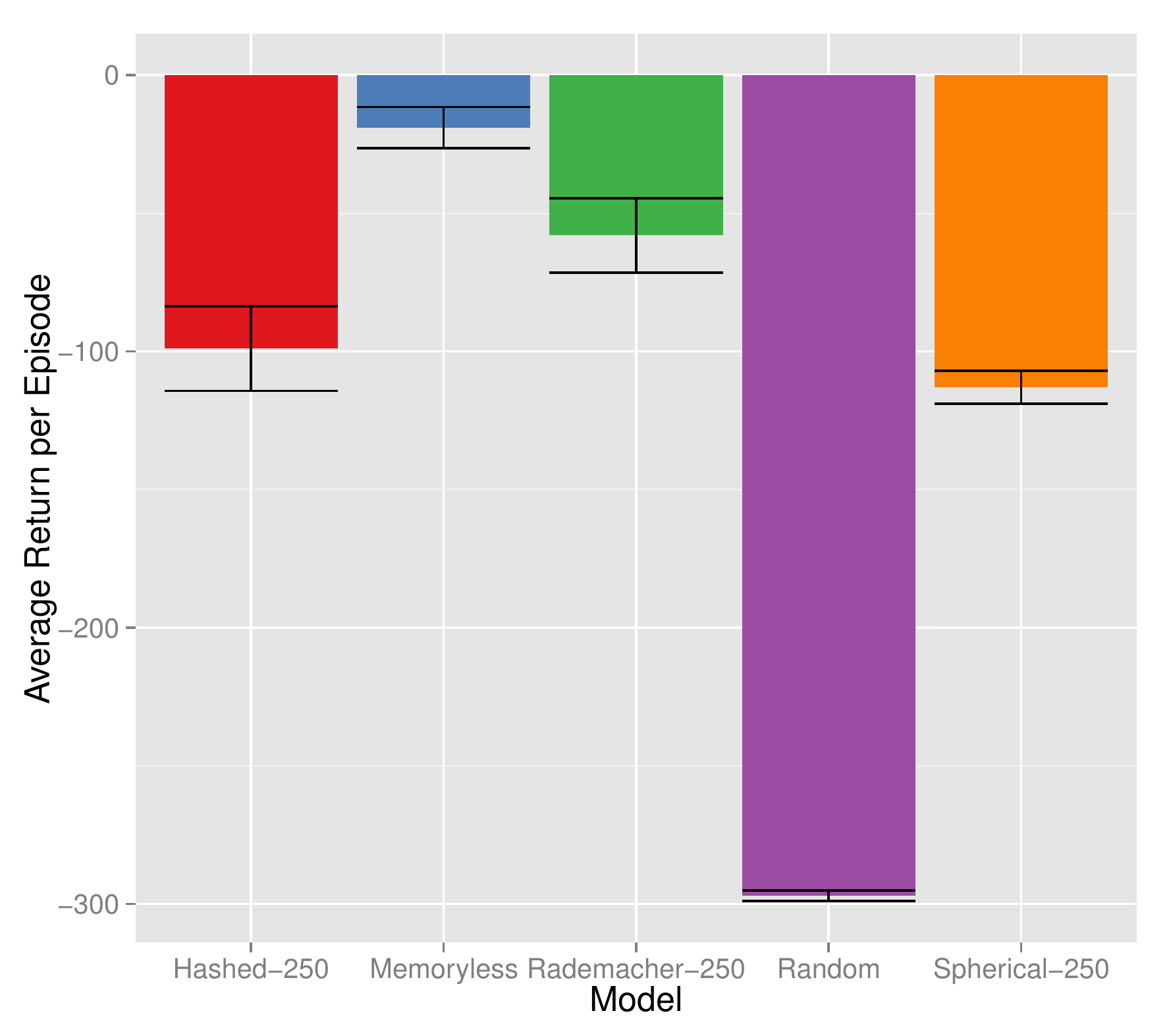}
                \caption{\textit{PocMan}}
        \end{subfigure}%
        ~ %add desired spacing between images, e. g. ~, \quad, \qquad etc.
        \begin{subfigure}[b]{0.5\textwidth}
                \includegraphics[height=7cm,width=7.5cm]{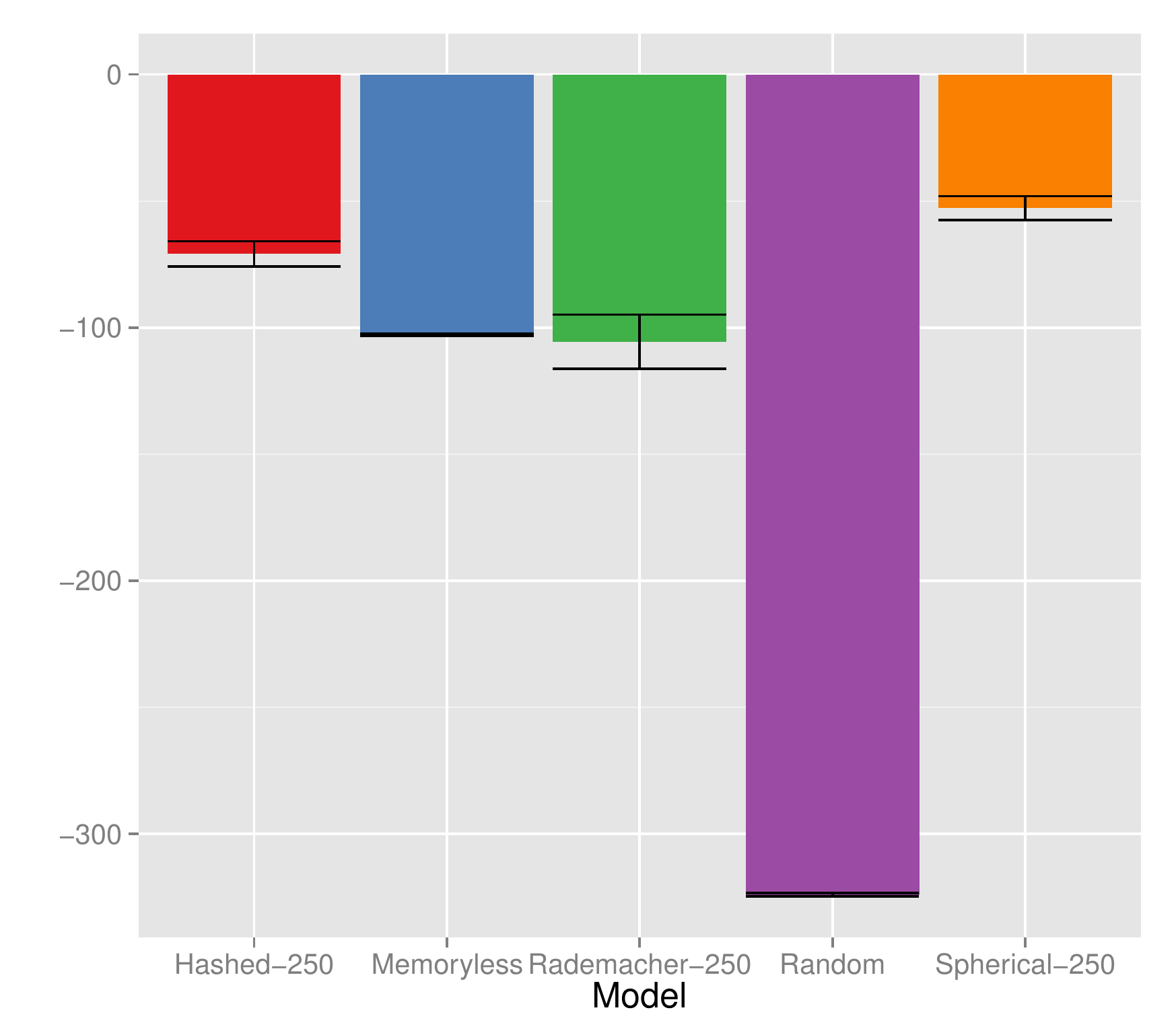}
                \caption{\textit{S-PocMan}}
        \end{subfigure}
        \caption{Average return per episode achieved in the \textit{PocMan} (a) and \textit{S-PocMan} (b) domains using different models and the baselines. Compressed dimension sizes are listed next to the model names. $95 \%$ confidence interval error bars are shown.}\label{fig:pacplan}
\end{figure}

For both \textit{PocMan} and \textit{S-PocMan}, we set $d'=25$ and examined compressed dimensions in the range $[250,500]$ (selecting only the top performer via a validation set); no TPSR models were used on these domains, as their construction exhausted the memory capacity of the machine used.
Following \cite{Veness:2011}, for these domains we use  $\gamma=0.99999$ as a discount factor.

Figure \ref{fig:pacplan} details the performance of the CPSR algorithms on the \textit{PocMan} and \textit{S-PocMan} domains.
In these domains, we see a much smaller performance gap between the CPSR approaches and the memoryless baseline.
In fact, in the \textit{PocMan} domain, the memoryless controller is the top-performer.
This demonstrates, first and foremost, that the \textit{PocMan} domain is not strongly partially observable.
Though the observations do not fully determine the agent's state, the immediate rewards available to an agent (with the exception of reward for eating the power pill and catching a ghost) are discernible through the observation vector (e.g., the agent can see locally where food is). 
Thus, the memoryless controller is able to formulate successful plans despite the fact that is treating the domain as if it were fully observable.
Moreover, a qualitative comparison with the Monte-Carlo AIXI approximation \citep{Veness:2011} reveals that the quality of the memoryless controller's plans are actually quite good.
In that work, they use a slightly different optimization criteria of optimizing for average transition reward, and with on the order of $50000$ transitions they achieve an average transition reward in the range $[-1,1]$ (depending on parameter settings).
With on the order of $250000$ transitions they achieve an average transition reward in the range $[1,1.5]$.
In this work, the memoryless controller achieves an average transition reward of $-0.2$ (despite the fact that it is actually optimizing for average return per episode), and it is thus, competitive given the same magnitude of samples, as approximately $50000$ transitions were used in this work.
It is also important to note that PSR-type models may be combined with memoryless controllers as memory PSRs (described in Section \ref{sec:related}), and so it should be possible to boost the performance of the CPSR models to match that of the memoryless controller in that way.

Importantly, in \textit{S-Pocman} where part of the observation vector is dropped and the rewards are sparsified, we see that the top-performer is again a CPSR based model (which in this case uses spherical projections).
This matches expectations since the food-rewards are no longer fully discernible from the observation vector, and thus the domain is significantly less observable. It is also worth noting that building naive TPSRs (without compression or domain-specific feature selection) is infeasible computationally in these PacMan-inspired domains, and thus the use of a PSR-based reinforcement learning agent (via the compression techniques used) in these domains is a considerable advancement.

A final observation is that the performance is quite sensitive to the choice of projection matrices in these results.
For example, in the \textit{S-PocMan} domain, the Rademacher projections perform no better than the memoryless baseline, whereas for \textit{PocMan} the Rademacher outperforms the other projection methods.
The exact cause of this performance change is unclear.
Nevertheless, this highlights the importance of evaluating different projection techniques when applying this algorithm in practice.

\subsubsection{Adaptive Migratory Management}

\begin{figure}
\centering
\includegraphics[scale=0.7]{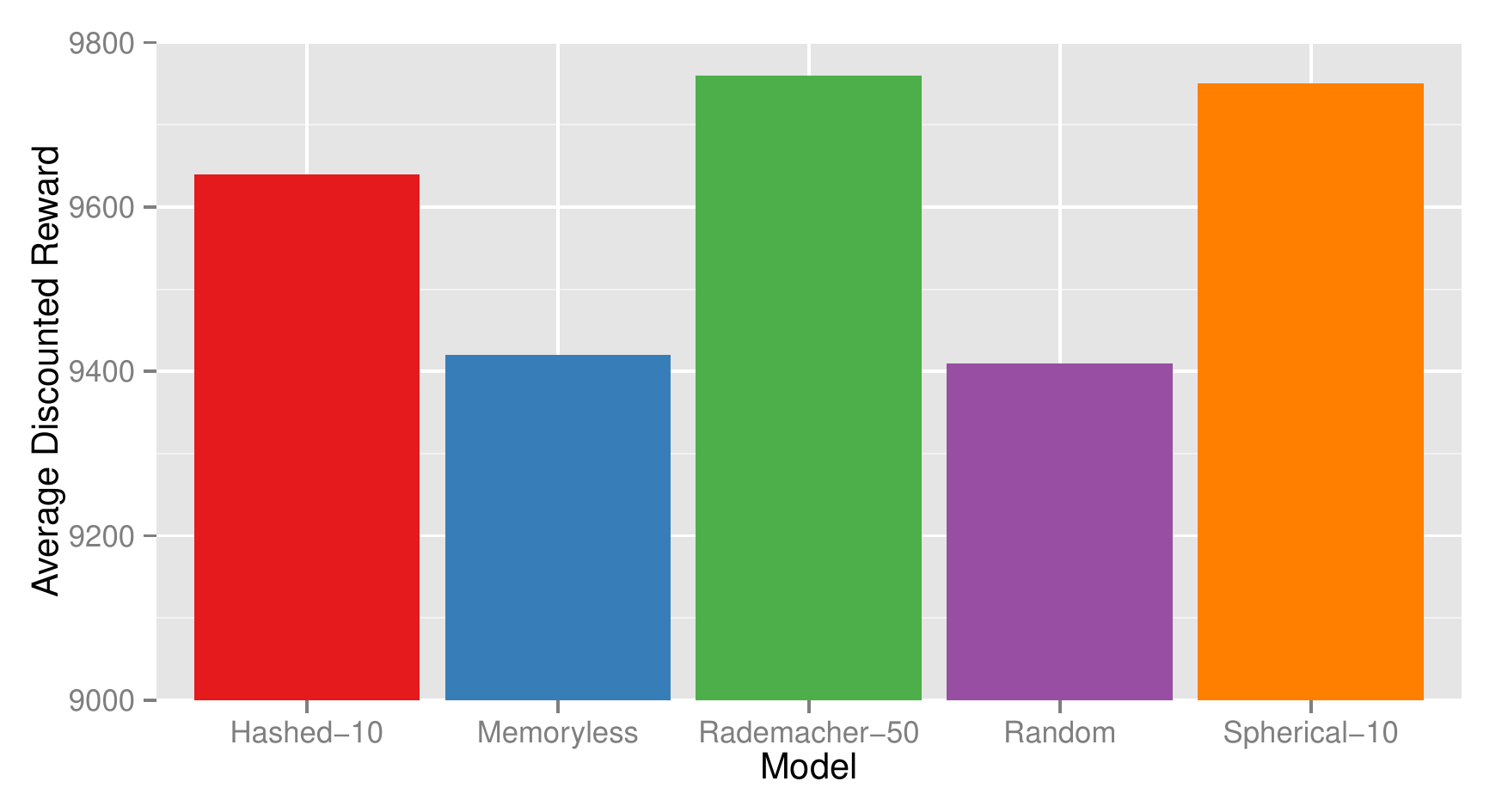}
\caption{Average discounted reward per episode (i.e., average return per episode) achieved in the \emph{AMM} domain using different methods over 100000 test episodes (each of length 50). The numbers beside the CPSR method names denote the projected dimension size. 95\% confidence intervals are too small to be visible.}\label{fig:discounted}
\end{figure}

\begin{figure}
\centering
\includegraphics[scale=0.7]{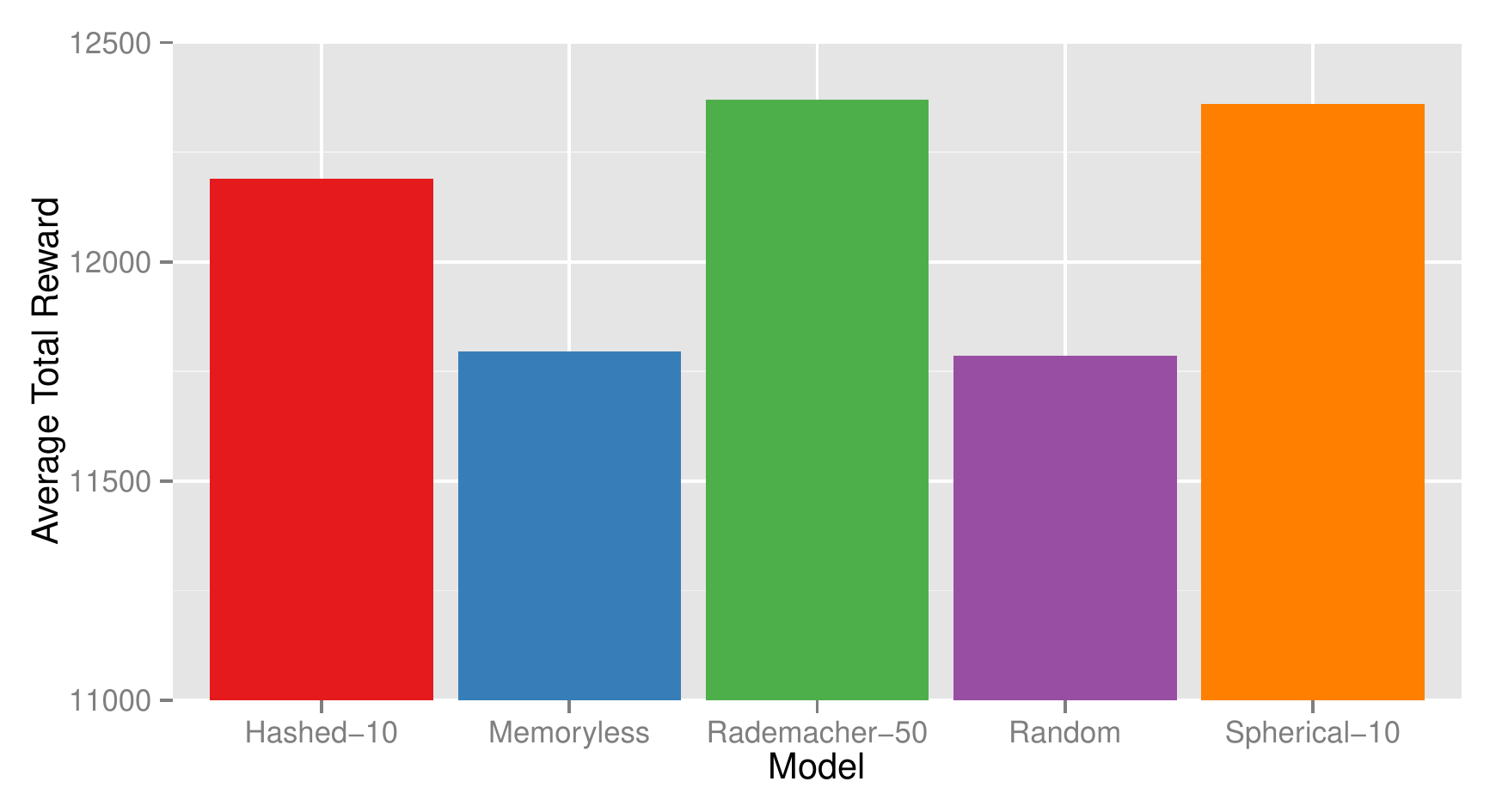}
\caption{Average total (undiscounted) reward per episode achieved in the \emph{AMM} domain using different methods over 100000 test episodes (each of length 50). The numbers beside the CPSR method names denote the projected dimension size. 95\% confidence intervals are too small to be visible.}\label{fig:average}
\end{figure}

We used a discount factor of $\gamma = 0.99$ for the \emph{AMM} domain.
For model-learning, we set $d'=10$ and examined compressed dimension in the range $[10,100]$.
The trajectories used during learning are all of maximum length 50 (the simulation may terminate earlier if all the birds perish). Note that since the \textit{AMM} domain is non-stationary \citep{Nicol:2013}, the model-learning algorithms must incorporate histories of length up to 50 (i.e., the entire trajectory) \citep{Boots:2009}, making the history dimension extremely large (i.e., $\approx 100000$) and making uncompressed PSR learning infeasible.
Tests up to length 4 were used for this task.

The results obtained are summarized in figures \ref{fig:discounted} and \ref{fig:average}.
Figure \ref{fig:discounted} shows the average sum of discounted rewards obtained using each method while figure \ref{fig:average} shows the average total (i.e., undiscounted) sum of rewards obtained by each method.
Both these test metrics are included as the discounted return is what the fitted-$Q$ algorithm optimizes for while the average total return is important from an intuitive perspective in that ecological conservationists do not necessarily discount the future. (Note, however, that the discount factor is necessary algorithmically for convergence since this domain technically has an infinite horizon).

Clearly, the CPSR methods are the top-performers with respect to both metrics.
In fact, the memoryless baseline does no better than random.
We also note that returns achieved by all methods are quite high.
The cause of the high return and the fact that the memoryless does no better than random are closely related.
Specifically, in the domain all actions are positive in that the agent must increase protection somewhere at each time-step. (The simulator does not allow for no action to be taken).
Thus, the random policy still leads to reasonable results since it will tend to spread its protection actions out uniformly randomly among the candidate nodes.
Moreover, without building a model and with access only to the observation vector at each time step, a reasonable strategy is to allocate protection to areas that have relatively low protection levels, compared to the other nodes.
That is, a reasonable memoryless strategy is also to simply spread out the protection among the candidate nodes, since without knowledge of the underlying dynamics one must assume that all nodes are equal.
Thus, intuitively the optimal memoryless strategy should be close to uniformly random, and this explains the similarity in scores between these two baselines.

Between the different CPSR methods, the Rademacher-projection based method performed the best with the spherical-projection method only performing slightly worse.
This result is expected in that there are stronger theoretical guarantees for these methods compared to the hashed projection method.

Lastly, we see that the results are consistent across the two metrics.
Interestingly, however, the performance increase between the top CPSR method and the random baseline is greater for the total (undiscounted) reward metric.
For that metric, the total reward obtained via the top-performing CPSR method is $4.6\%$ greater than the baseline, whereas for the discounted metric the top-performing method scores $3.7\%$ greater than the random baseline.
This makes sense in that the CPSR models should benefit more at longer horizons, since (1) it takes time for the CPSR model to incorporate observed information into its predictions and (2) the non-stationary in the domain, which is captured via the CPSR model, is only a factor at longer time-scales \citep{Nicol:2013}.

\section{Discussion}\label{sec:discuss}

The CPSR approach provides a new avenue for model-based reinforcement learning where agents must formulate policies in large, complex partially observable domains without access to a fully-specified prior system model (i.e., where the system model must be learned prior to planning).
The compressed learning algorithm allows accurate approximations of PSR models to be constructed in a memory and time efficient manner, and the use of random projections regularizes the learned solutions, preventing high variance models (over-fitting) and potentially leading to more accurate results.
We elucidated theoretical guarantees bounding the induced approximation error of this model-learning approach, showing that the low-dimensional embeddings of the models retain predictive accuracy.
In addition, we proposed a planning approach which exploits these compressed models in a principled manner, allowing for high-quality plans to be constructed without prior domain knowledge.
Finally, we outlined how model-learning and planning can be combined at a high-level.

The empirical results we obtained demonstrate the efficacy of this approach and delineate domains in which its use is beneficial.
The model quality experiments demonstrate that CPSR models achieve predictive accuracy competitive to that of uncompressed models, while taking a fraction of the runtime, and the planning results demonstrate that these models can be exploited by efficient planners, providing a novel and powerful framework for model-based reinforcement learning. Moreover, the results highlight the fact that the benefits of such a model-based approach are most stark in domains that are not only partially observable in the traditional sense but that are also strongly partially observable in that the $Q$-function (or a good approximation of it) is not discernible from the observation vectors.
In other words, the results demonstrate that aliased observations (and an unobserved hidden state) alone do not necessitate the need for a model-based learning algorithm. 
A model-based approach only becomes necessary when the observations are not sufficient for learning a reasonable approximation of the $Q$-function.

\subsection{Practical Concerns}

The implementation of complex RL frameworks often reveals practical issues that are not immediately apparent given formal descriptions.
In order to facilitate the use of the CPSR algorithm in applications, we outline some pertinent practical issues that arise while implementing the CPSR algorithm and describe our solutions.

\subsubsection{Selecting the Projection Matrices}

First, it is necessary to reiterate the sensitivity of the approach with respect to both the projection dimension and type of projection used.
Empirically, we found that the results could be quite sensitive to these parameters, though this was only the case for some domains.
For example, selecting a projection dimension that is too small may lead to suboptimal (near-random) performance.
This issue is further exacerbated by the fact that the true dimension of the underlying system is unknown.

The cause of the sensitivity with respect to the projection size is quite evident (smaller dimensions lose information but provide more regularization). However, the underlying  cause of the differing performance between the different projection types is not as clear.
One would expect the hash-type projection matrices to perform differently than the Rademacher and spherical projections, since the hash-type matrices do not satisfy the JL lemma, but we witnessed substantial variation between all three projection types, especially on the PacMan-type planning domains.
Moreover, for the \emph{ColoredGridWorld} domain, the difference between the projection types was more stark for planning performance compared to prediction performance.

The results thus indicate that  planning performance is more sensitive to the choice of the projection matrix (compared to prediction performance). 
One explanation for this is simply that small discrepancies in the prediction performance of the models are amplified when agents must plan using the predictive models.
The differing results obtained using the different projection matrices may then be due to the fact that a coarse-grained search (necessitated by computational requirements) for the compressed dimension-size was used and that different random projections may be optimal for slightly different projection sizes \citep{Achlioptas:2001}.
For example, a Rademacher projection may be near optimal at one point on the coarse-grained search while a spherical projection may be optimized at a point not included in the coarse-grained search.
The slight differences in model-quality induced by the coarse-grained search would then propagate and lead to large variations in planning performance.

In order to cope with the sensitivity of the CPSR approach with respect to the projection sizes and dimension, we recommend using multiple phases of grid search (starting with exponentially separated values).
Moreover, it is useful to narrow down the size-range for the projections using model-quality experiments (before performing hyperparameter optimization for planning), since model-quality experiments are not as computational expensive (compared to planning experiments).

\subsubsection{Improving Efficiency by Caching}

In Section \ref{sec:learningcpsrs} we defined the projection operators via the functions $\phi_\TT \::\: \TT \rightarrow \mathbb{R}^{d_\TT}$ and $\phi_\HH \::\: \HH \rightarrow \mathbb{R}^{d_\HH}$. This specification engenders a number of benefits.
Specifically, the full projection matrices do not need to be held in memory and the number of tests and histories do not need to be specified in advance.
There is a runtime penalty associated with the technique, however, as the mappings must be recomputed each time a particular test or history is encountered while iterating over the sample trajectories.
In order to ameliorate this issue, while retaining the benefits of specifying the projections as functions, we implemented a least-recently-used (LRU) cache.
By caching the mappings for frequently encountered tests and histories, we improved the empirical runtime of the algorithm considerably. 

\subsubsection{Numerical Stability Issues}\label{sec:stability}

At its core, the CPSR algorithm relies on standard linear algebra techniques, namely SVD and matrix inversions, which are prone to numerical stability issues.
If the matrices upon which these operations are performed are ill-formed, suboptimal results will be obtained (or the algorithm will simply fail).
In this work, we found one common situation where such stability issues arise.

Since we do not normalize the probability estimates in Section \ref{sec:learningcpsrs}, the singular values of $\cPthe$ in \eqref{eq:svd} grow with the size of the training set.
This leads to stability issues when inverting the matrix of singular values in order to compute the implicit pseudoinverse  in \eqref{eq:cinfdef} and \eqref{eq:Caodef}.
This stability issue can be alleviated by normalizing the probability estimates, or more generally, by scaling $\cPthe$ by a small constant.
Since this constant cancels out during learning, it can be picked arbitrarily, but it should be chosen such that the magnitude of the values in $\cPthe$ are near unity.
The most straightforward approach is to simply normalize the probability estimates, though this may not always suffice (e.g.,  if there are extremely unlikely events, the normalizer may make certain entries too small leading to further stability issues).
We also empirically observed that setting $d' < d_{\TT}$ and/or removing singular values below a certain threshold (a standard technique) helped with numerical stability. 

%We also observed significant stability issues when performing  goal-directed sampling with large sample sizes.
%Goal-directed sampling induces a bias in the observed distribution of test/history pairs, and most frequently this bias leads to certain test/history pairs being encountered more than others. 
%Thus if large sample sizes are used with goal-directed sampling, the $\cPthe$ matrix becomes imbalanced in that certain entries dominate in terms of magnitude.
%As a result of this imbalance, the SVD in \eqref{eq:svd} is again unstable, and poor results are obtained.
%In fact, we found that performing more than 4 iterations of goal-directed sampling in the \textit{ColoredGridWorld} domain (using 50 trajectories per iteration) degraded performance to no better than random.
%In this case, there is no remedy beyond using small sample sizes or random sampling.
%However, we obtained the best results using random sampling and that method is statistically unbiased (while goal-directed sampling is biased), so this problem is not of great import. 

\subsubsection{$Q$-function Approximation and Sampling Strategies}

Algorithm 2 in Section \ref{sec:planning} permits a wide-variety of sampling strategies, and the sampling strategy used implicitly constrains the $Q$-function approximation obtained.
In this work, we used an unbiased random sampling strategy in the batch setting.
That is, we collected a large batch of random samples, which we used to both learn a model and construct plans.
We opted for this framework as (1) our simulators were designed for the batch setting and (2) the theoretical results of Section \ref{sec:theory} assume a blind (random) sampling strategy is used.

We did, however, experiment with a goal-directed sampling approach \citep{Ong:2012}, where phases of exploration and exploitation are interleaved.
In the goal-directed paradigm, a number of mini-batch sampling iterations are used, and the sampling policy ($\pi_s$) is updated at each iteration to be $\epsilon$-greedy over the agent's current policy ($\pi_i$).
\cite{Ong:2012} found that this approach led to better performance in the small-sample setting.
In our experiments, where we used larger numbers of samples (on the order of $10000$), we found that the goal-directed approach did not improve over random sampling and, in fact, often led to worse results and numerical instabilities.
In particular, the bias in the sampling strategy led to an imbalance in the $\cPthe$ matrix in that certain entries dominated in terms of magnitude.
As a result of this imbalance, the SVD in \eqref{eq:svd} became unstable, and poor results were obtained.
Such stability problems are likely to be an issue whenever biased sampling strategies are used in the large-sample batch setting.
However, in online or small sample settings, such strategies will likely lead to performance increases due to the fact that their exploration is myopic and focusses on areas of the state-space relevant to planning \citep[as shown by][]{Ong:2012}.

\subsubsection{Compressing Histories}

The theoretical analysis of Section \ref{sec:theory} assumes that $\PhiH$ has orthonormal columns.
However, in order to obtain maximal computational benefits, it is necessary to use a compressive $\PhiH$, i.e. a $\PhiH$ that acts as a feature selector on histories.
In fact, for massive domains such as the PacMan-style domains, compressing histories is necessary for tractable learning and planning.

Viewing CPSR learning from the perspective of regression (as was done throughout this paper), the compression of histories is equivalent to compressing the samples used for regression; that is, it is equivalent to linearly mixing the samples. 
More formally, we use the transformation
\begin{align}
\mb{y} = \mb{X}^\top\mb{w} + \boldsymbol{\eta}
\rightarrow 
\PhiH\mb{y} = \PhiH\mb{X}^\top\mb{w} + \PhiH\boldsymbol{\eta},
\end{align}
where as usual $\mb{X}$ is a design matrix, $\mb{w}$ a vector of regression weights, $\mb{y}$ a vector of targets, and $\boldsymbol{\eta}$ a vector of noise terms.
Intuitively, we can view this projection by $\PhiH$ as roughly averaging over training samples.
The number of samples for the regression will then be reduced, but the averaged samples will have reduced (maximum) variance in their noise terms.

Of course, in this work, we use random $\PhiH$ matrices, which do not correspond directly to taking averages over samples.
The most important implication of this is that the noise terms of the new combined samples are not independent. 
This more complex setting has been analysed in detail by \cite{Zhou:2007} (for random Gaussian matrices).
In that work, they focus on the more specific setting of $l_1$ regularized regression, and they prove a number of important results.
Of particular relevance is 
Claim 4.3, which shows (under certain conditions) that the entry-wise discrepancy between $\mb{Q}^\top\mb{Q}$ and $\mb{Q}^\top\mb{\Phi}^\top\mb{\Phi}{\mb{Q}}$ decreases asymptotically to zero almost surely, where $\mb{Q} \in \R^{n \times m}$ and $\mb{\Phi} \in \R^{d \times n}$ is a random Gaussian matrix defined as in Theorem \ref{theorem:cao-error}. 
This key result facilitates bounding the discrepancy between the compressed training error and the true error of the regressor and does not rely on $l_1$ regularization assumptions.
We refer the interested reader to that work for detailed proofs.

Finally, we reiterate that in this work the compression of histories is a computational necessity, as it allows us to scale the learning algorithm to domains that would be intractable otherwise.
And empirical investigations in Section \ref{sec:empirical} show that the compression of histories to $d_{\HH}=d_{\TT}$ introduces only a small amount of error during model-learning.

\subsection{Related Work}\label{sec:related}

The CPSR algorithm is closely related to work on using features or kernel embeddings with PSRs \citep{Boots:2009,Boots:2011,Boots:2013b}, where features of tests, histories, and/or observations are employed.
Indeed, one view of the CPSR learning approach is that it is an instantiation of the feature-based learning approach where principled random features are employed.
However, this view is limited in the sense that the random features used here facilitate an analysis in terms of compression, whereas with other feature-based PSR methods it is simply assumed that the specified features are sufficient to capture the structure of $\Pth$; that is, the standard feature-based methods assume features that are not compressive \citep{Boots:2009,Boots:2011,Boots:2013b}.

This distinction of whether or not features are assumed as compressive also highlights the differing motivations between existing feature-based PSR learning and the CPSR approach: in the CPSR approach, compressive random features are employed to increase the efficiency and scalability of learning, whereas in other works \cite[e.g.][]{Boots:2009,Boots:2011,Boots:2013b} the features are used to facilitate learning in domains with continuous or structured observation spaces.

It should be noted, however, that since the general PSR learning framework assumes discrete observations, decomposing a continuous domain via feature extraction is necessary for learning in that setting.
Moreover, \cite{Boots:2013b} shows how the well-known ``kernel trick'' can be employed to learn in feature-spaces of infinite dimension.
The penalty associated with this kernel embedded approach is that learning scales cubically with the number of training examples, leading to high computational overhead \citep{Boots:2013b}.
\cite{Boots:2011} show how to partially alleviate this cost by using random features to approximate certain kernels, a technique that also relies on random projections (though not in the compressed sensing setting).

%An interesting open question is how random projections, or related techniques, can be combined with these feature-based methods, allowing for efficient learning in continuous domains.
%The CPSR learning approach could certainly be combined with standard feature-based learning (though it is unclear how sparse these feature spaces are, which impacts our theoretical analysis).
%However, the optimal method for integrating random projections with a feature-based approach that uses the ``kernel trick'' is not obvious.

In a similar vein, the CPSR-based planner is closely related to the goal-directed planning and learning approach of \cite{Ong:2012}.
The primary difference between our work and this goal-directed approach is that we present a more general combined learning and planning framework, which accommodates the use of a wide variety of sampling strategies. 

Beyond these works, our approach bears similarities to the memory PSR (mPSR) approach of \cite{James:2005}, which uses a type of hybrid PSR-MDP model to reduce computational costs and increase predictive accuracy, and the hierarchical PSRs (HPSRs) of \cite{Wolfe:2006}, which use the option framework \citep{Sutton:1999} to increase the predictive capacity of PSRs. 
Importantly, the improvements suggested by both these approaches are not incompatible with our compressed learning algorithm.

Our approach also shares similarities with certain model-based reinforcement learning algorithms, which use adaptive history-based techniques.
Examples of these algorithms include \emph{U-Tree} \citep{Mccallum:1996} and the \emph{Monte-Carlo AIXI approximation} \citep{Veness:2011}.
These approaches share the motivation of developing agents that can learn a model of dynamical system and plan using this model.
They differ, however, in the instantiation of their model-based approach, as they use an adaptive history-based approach, which intuitively corresponds to learning mixtures of different $k$-order MDPs (where $k$ varies adaptively).
A key aspect of these approaches is focussing the model-learning on areas of the state-space relevant to achieving goals (similar to the goal-directed sampling routine) \citep{Mccallum:1996,Veness:2011}. 
Thus, a fundamental difference between Monte-Carlo AIXI-like approaches and the one proposed here is that they efficiently learn myopic models, necessarily constrained by the planning aspect of the problem, whereas in this work we retain the option of learning full-unbiased models of domains (i.e., our model-learning may be decoupled from planning).
One implication of this is that the models learned via the CPSR learning approach may be reused in different planning contexts.
However, a disadvantage of learning  complete (i.e., full and unbiased) models is that it can be impractical in very large and complex domains.

\subsection{Future Directions}

Given the above discussion, an interesting direction for future work would be an analysis of the inductive bias associated with both the PSR and Monte-Carlo AIXI paradigms.
Though these methods bear similarities, their theoretical motivations are quite distinct: PSRs being motivated by the theory of observable operators while certain AIXI-like methods have information-theoretic (and/or Bayesian) motivations \citep{Veness:2011}.
Recently, there have been a number of theoretical advancements in the understanding of observable operator methods, such as the local loss formulation of \cite{Balle:2012} and the method of moments formulation of \cite{Anandkumar:2012}.
These advancements could serve as tools in such an analysis.
Perhaps the most interesting question in this area is understanding the regularization induced by these different paradigms (e.g., due to the restriction of the model classes).
For example, the Monte-Carlo AIXI method explicitly penalizes model complexity, while this does not explicitly factor into the optimization of PSR-type methods (besides through the hyper-parameter selection of the model-size).

Another interesting avenue for the continuation of this work is exploring the use of different optimization frameworks during learning.
In this work, we implicitly use the standard least-squares objective when solving the pseudoinverse in \eqref{eq:cinfdef} and \eqref{eq:Caodef}.
However, there is no a priori reason to believe that this is the optimal formulation, and in fact, promising results have been obtained by modifying this optimization (e.g., through convex-relaxation) \citep{Balle:2012}.
Moreover, it is possible that alternative formulations may reveal novel regularization strategies (e.g., regularization on the implicit observable-operator structure) and additional algorithmic efficiencies.

Lastly, the framework presented here provides the necessary ingredients for applying a CPSR-based learning and planning framework to difficult real-world application problems, such as robot navigation problems similar to those solved by U-tree-based approaches \citep{Mccallum:1996}.
Of course, such applications would introduce certain engineering issues not highlighted here.
In particular, the sampling strategy, projection size, and projection type would necessarily be constrained by the problem domain and by hardware limitations; for example, it may be worthwhile to use highly optimized Rademacher projections.
Moreover, in domains with extremely large action and observation dimensions, using a distributed implementation (e.g., of equation \eqref{eq:Caodef} in the learning algorithm) would likely engender significant computational benefits.
And, in domains with continuous observations, it would be necessary to combine discretization or kernel-based feature extraction with the CPSR compression techniques. 
These engineering issues, however, should not necessitate altering the core of the CPSR approach.

% Acknowledgements should go at the end, before appendices and references
\acks{The authors would like to thank Doina Precup, Yuri Grinberg, Sylvie Ong, and Clement Gehring for helpful discussions on this work, and David Silver and Joel Veness for support on the PocMan domain. We are very grateful to our anonymous reviewers for their comments and recommendations.  Financial support for this work was provided by NSERC Discovery and CGS-M grants.}

\appendix
\section*{Appendix A.}
\subsection*{A.1 Proof of Theorem~\ref{theorem:cao-error}}
\begin{proof}
With eigenvalue decomposition we have $E_{\rho(h)}[\Pqohh \Pqohh^\top] = \mb V \mb D \mb V^\top$, where $\mb D$ is the diagonal matrix containing the eigenvalues and $\mb V$ is an orthonormal basis.
 Let $\mb I_m$ be a $|\QQ| \times |\QQ|$ matrix with the first $m$ diagonal elements set to 1 and 0 elsewhere. 
 For all $1 \leq i \leq d$, define: 
 $[\tilde {\mb{\Phi}}]_{i,*}= [\mb{\Phi}]_{i,*} \mb V \mb I_m \mb V^\top$ and $[\mb{\Phi}']_{i,*} = [\mb{\Phi}]_{i,*} \mb V$. 
 Note that since $\mb V$ is an orthonormal basis and $[\mb{\Phi}]_{i,*}$ is i.i.d. normal, $[\mb{\Phi}']_{i,*}$ will also have an i.i.d. normal distribution with the same covariance.

We wish to substitute $[\mb{\Phi}]_{i,*}$ with $[\tilde {\mb{\Phi}}]_{i,*}$ which has a small norm and introduces a small bias. We first bound the norm of $[\tilde {\mb{\Phi}}]_{i,*}$ as follows. With probability no less than $1-\delta/4$ for all $1 \leq i \leq d$:
\begin{eqnarray}
\|\tilde{[\mb{\Phi}}]_{i,*}\|^2 &=& [\mb{\Phi}]_{i,*} \mb V \mb I_m \mb V^\top \mb V \mb I_m \mb V^\top  [\mb{\Phi}]_{i,*}^\top\\
&=& [\mb{\Phi}']_{i,*} \mb I_m ([\mb{\Phi}']_{i,*})^\top = \sum^{m}_{j=1} ([\mb{\Phi}']_{ij})^2\\
&\leq& m + 4 \sqrt{m} \ln (4d/\delta).
\end{eqnarray}
The tail bound in last line is union bounding over a corollary of Lemma 1 in \cite{laurent2000adaptive}. The bias induced by using $\tilde {[\mb{\Phi}}]_{i,*}$ can be bounded as well. Define $b(h) = [\mb{\Phi}]_{i,*} \Pqohh - [\tilde {\mb{\Phi}}]_{i,*}\Pqohh$. With probability no less than $1-\delta/4$ for all $1 \leq i \leq d$:
\begin{eqnarray}
\|b(h)\|^2_{\rho(h)}
&=& E_{\rho(h)}[ ([\mb{\Phi}]_{i,*} - [\tilde {\mb{\Phi}}]_{i,*}) \Pqohh  \Pqohh^\top ([\mb{\Phi}]_{i,*} - [\tilde {\mb{\Phi}}]_{i,*})^\top] \\
&=& ([\mb{\Phi}]_{i,*} -[\tilde {\mb{\Phi}}]_{i,*}) \mb V \mb D \mb V^\top ([\mb{\Phi}]_{i,*} - [\tilde {\mb{\Phi}}]_{i,*})^\top \\
&=& ([\mb{\Phi}]_{i,*} -[\mb{\Phi}]_{i,*} \mb V \mb I_m \mb V^\top) \mb V \mb D \mb V^\top ([\mb{\Phi}]_{i,*} - [\mb{\Phi}]_{i,*} \mb V \mb I_m \mb V^\top)^\top \\
&=& [\mb{\Phi}]_{i,*} \mb V \mb (\mb I - \mb I_m) \mb D (\mb I - \mb I_m) \mb V^\top [\mb{\Phi}]_{i,*}^\top \\
&=& [\mb{\Phi}']_{i,*} \mb (\mb I - \mb I_m) \mb D (\mb I - \mb I_m) ([\mb{\Phi}']_{i,*})^\top \\
&=& \sum^{|\QQ|}_{j=m+1} ([\mb{\Phi}']_{ij})^2 \sigma^2_j\\
&\leq& \nu + 4 \sqrt{\nu} \ln (4d/\delta).
\end{eqnarray}
The tail bound again is due to Lemma 1 in \cite{laurent2000adaptive} using the assumption $\sigma^2_m \leq 1$. Using the above bounds, we have for for all $1 \leq i \leq d$:
\begin{align}
\forall h: [\mb{\Phi}]_{i,*} \Pqohh = [\tilde {\mb{\Phi}}]_{i,*}\Pqohh + b(h) = ([\tilde {\mb{\Phi}}]_{i,*}\Bao) \Pqhh + b(h).
\end{align}
Therefore, we have a target $[\mb{\Phi}]_{i,*} \Pqohh$ that is near-linear in the sparse features $\Pqhh$, with expected bias bounded by $b^2 = \nu + 4 \sqrt \nu \ln(4d/\delta)$, and norm of the weight vector $[\tilde {\mb{\Phi}}]_{i,*}\Bao$ bounded by $w^2 = \|\Bao\|^2(m + 4 \sqrt{m} \ln (4d/\delta))$.

%w^2 &=& \|\Bao\|^2(m + 4 \sqrt{m} \ln (4d/\delta)),\\
%x^2 &=& \|\Pqhh\|^2_{\rho(h)},\\
%b^2 &=& \nu + 4 \sqrt \nu \ln(4d/\delta),\\
%\sigma_\eta^2 &=& \frac{4 k \ln(4|\QQ|/\delta)}{d} \sigma^2_y + w^2 \sigma^2_x.

By definition, $\mb u_i$ is the COLS estimate with input $\ePqh$, target $[\mb{\Phi}]_{i,*} \ePqoh$, and projection $[\mb{\Phi}]_{-i,*}$. But in order to use the bound of Equation~\ref{eqn:colsbound}, we need to find the corresponding noise parameters of the COLS algorithm. Since, unlike the assumption of the general COLS bound, both the input and the output of the regression are noisy, we need to derive the effective overall noise variance in the sample output. We have:
\begin{eqnarray}
[\mb{\Phi}]_{i,*} \ePqohh &=& [\mb{\Phi}]_{i,*} \Pqohh + [\mb{\Phi}]_{i,*} \Delta_y\\
&=& [\tilde {\mb{\Phi}}]_{i,*}\Pqohh + b(h) + [\mb{\Phi}]_{i,*} \Delta_y\\
&=& [\tilde {\mb{\Phi}}]_{i,*}\Bao (\ePqhh - \Delta_x) + b(h) + [\mb{\Phi}]_{i,*} \Delta_y\\
&=& ([\tilde {\mb{\Phi}}]_{i,*}\Bao) \ePqhh + b(h) + ([\mb{\Phi}]_{i,*} \Delta_y - [\tilde {\mb{\Phi}}]_{i,*}\Bao \Delta_x).
\end{eqnarray}
And thus the sample points are:
\begin{align}
\ePqhh \rightarrow ([\tilde {\mb{\Phi}}]_{i,*}\Bao) \ePqhh + b(h) + ([\mb{\Phi}]_{i,*} \Delta_y - [\tilde {\mb{\Phi}}]_{i,*}\Bao \Delta_x).
\end{align}
The effective noise $[\mb{\Phi}]_{i,*} \Delta_y - [\tilde {\mb{\Phi}}]_{i,*}\Bao \Delta_x$ has mean 0. Since $\Delta_y$ is $k$-sparse and $\|[\tilde {\mb{\Phi}}]_{i,*}\Bao\|^2\leq w^2$, the variance of the effective noise term is bounded by $\max_j ([\mb{\Phi}]_{ij})^2 k \sigma^2_y + w^2 \sigma^2_x$. Maximization over $i$ and using a tail bound on the maximum of squared normals gives the $\sigma_\eta^2$ defined in the theorem.

We now apply the union bound to Equation~\ref{eqn:colsbound}. With probability no less than $1-\delta/4$, for all  $1 \leq i \leq d$:
\begin{align}
\left\| \mb u_i ([\mb{\Phi}]_{-i,*}\Pqhh)  - [\mb{\Phi}]_{i,*} \Pqohh \right\|_{\rho(h)}
\leq \epsilon(|\HH|, |\QQ|, d, w^2, x^2, b^2, \sigma_\eta^2, \delta/4d).
\end{align}
Note that by our definition of $\Cao$, we have that $\mb u_i ([\mb{\Phi}]_{-i,*}\Pqhh) =  (\Cao)_i (\mb{\Phi}\Pqhh)$, which immediately gives the theorem by combining the error bounds on each row.
\end{proof}

\subsection*{A.2 Proof of Theorem~\ref{theorem:cinf-error}}
\begin{proof}
Similar to Theorem~\ref{theorem:cao-error}, we have $\Phh = \binf^\top \Pqhh$ for all $h$.
Therefore we have a linear target and by definition $\cinf$ is the COLS estimate with projection $\mb{\Phi}$. We have:
\begin{eqnarray}
\ePhh &=& \Phh + \Delta_z = \binf^\top \Pqhh + \Delta_z \\
&=& \binf^\top \ePqhh - \binf^\top \Delta_x + \Delta_z.
\end{eqnarray}
Thus the effective variance is bounded by the $\sigma^2_\infty$ defined in the theorem. We complete the proof by an application of the bound in Equation~\ref{eqn:colsbound}.
\end{proof}

\subsection*{A.3 Proof of Theorem~\ref{theorem:prop-error}}
\begin{proof}
For all $t$, define $\mb e_t = \mb C_{a_{t}o_{t}} \mb C_{a_{t-1}o_{t-1}} \dots \mb C_{a_{1}o_{1}} \mb c_1 - \Pqaot$. After applying the $n$th compressed operator we have:
\begin{eqnarray}
\| \mb e_n \|_{\rho_{n}} &=&  \| \mb C_{a_{n}o_{n}} \mb C_{a_{n-1}o_{n-1}} \dots \mb C_{a_{1}o_{1}} \mb c_1 - \Pqaon \|_{\rho_{n}} \\
&=&  \| \mb C_{a_{n}o_{n}} (\Pqaonm + \mb e_{n-1}) - \Pqaon \|_{\rho_{n}} \\
&\leq&  \| \mb C_{a_{n}o_{n}} \mb e_{n-1}\|_{\rho_{n}} + \| \mb C_{a_{n}o_{n}} \Pqaonm - \Pqaon \|_{\rho_{n}} \\
&\leq&  \| \mb C_{a_{n}o_{n}} \mb e_{n-1}\|_{\rho_{n}} + \max_{o_n, a_n} \| \mb C_{a_{n}o_{n}} \Pqaonm - \Pqaon \|_{\rho_{n-1}} \\
&\leq&  c \|\mb e_{n-1}\|_{\rho_{n}} + \max_{o_n, a_n} \| \mb C_{a_{n}o_{n}} s_{n-1} \Pqaonm - s_{n-1} \Pqaon \|_{\rho} \label{line:scaled}\\
&\leq&  c \|\mb e_{n-1}\|_{\rho_{n-1}} + s_{n-1} \epsilon \\
&\leq& \epsilon \sum^{n-1}_{t=1} s_t c^{n-i-1}. \label{line:induction}
\end{eqnarray}
Line~\ref{line:scaled} uses the distribution assumption on $\rho_{n-1}$ and having $\Pqaon$ linear in $\Pqaonm$. Line~\ref{line:induction} follows by induction. We now apply the normalizer operator:
\begin{align}
\| \cinf \mb C_{a_{n}o_{n}} & \mb C_{a_{n-1}o_{n-1}} \dots \mb C_{a_{1}o_{1}} \mb c_1 - \PP(o_{1:n} || a_{1:n}) \|_{\rho_n} \\
&=  \| \cinf (\Pqaon + \mb e_{n}) - \PP(o_{1:n}  || a_{1:n}) \|_{\rho_{n}}\\
&\leq  \| \cinf \mb e_{n}\|_{\rho_{n}} + \| \cinf \Pqaon - \PP(o_{1:n} | a_{1:n}) \|_{\rho_{n}}\\
&\leq  \| \cinf \| \| \mb e_{n}\|_{\rho_{n}} + \| \cinf s_n \Pqaon - s_n \PP(o_{1:n} || a_{1:n}) \|_{\rho} \label{line:scaled1} \\
&\leq \|\cinf\| \epsilon \sum^{n-1}_{t=1} s_t c^{n-t-1} + \epsilon_\infty s_n. \label{line:cinferr}
\end{align}
Line~\ref{line:scaled1} uses the distribution assumption on $\rho_n$ and Line~\ref{line:cinferr} uses the bound of Theorem~\ref{theorem:cinf-error}.
\end{proof}

\vskip 0.2in
\bibliography{cpsr}

\end{document}